\documentclass{article}

    \PassOptionsToPackage{numbers, compress}{natbib}
\usepackage[preprint]{neurips_2024}

\usepackage[utf8]{inputenc} % allow utf-8 input
\usepackage[T1]{fontenc}    % use 8-bit T1 fonts
\usepackage{hyperref}       % hyperlinks
\usepackage{url}            % simple URL typesetting
\usepackage{booktabs}       % professional-quality tables
\usepackage{amsfonts}       % blackboard math symbols
\usepackage{nicefrac}       % compact symbols for 1/2, etc.
\usepackage{microtype}      % microtypography
\usepackage{graphicx}
\usepackage{amssymb}
\usepackage{algorithmic}
\usepackage{algorithm}
\usepackage{graphicx}
\usepackage{subcaption}
\usepackage{enumerate}
\usepackage{wrapfig}

\usepackage[table,xcdraw]{xcolor}
\usepackage{multirow,booktabs}
\usepackage[frozencache,cachedir=.]{minted}
\usepackage{lipsum}
\title{
Hammer: Robust Function-Calling for On-Device \\ Language Models via Function Masking
}

\author{%
  Qiqiang Lin$^{*,1}$\\
  \And
  Muning Wen$^{*,2}$\\
  \And
  Qiuying Peng$^{*,+,1}$\\
  \And
  Guanyu Nie$^{3}$\\
  \And
  Junwei Liao$^{2}$\\
  \And
  Jun Wang$^{1}$\\
  \And
  Xiaoyun Mo$^{1}$\\
  \And
  Jiamu Zhou$^{1}$\\
  \And
  Cheng Cheng$^{1}$\\
  \And
  Yin Zhao$^{1}$\\
  \And
  Jun Wang$^{+,1}$\\
  \And
  Weinan Zhang$^{+,2}$\\
}

\begin{document}

\maketitle

\let\thefootnote\relax\footnotetext{\footnotesize $^{*}$Equal Contribution, $^{+}$Corresponding Author, $^{1}$OPPO Research Institute, $^{2}$Shanghai Jiao Tong University, $^{3}$Iowa State University}

\begin{abstract}
Large language models have demonstrated impressive value in performing as autonomous agents when equipped with external tools and API calls. Nonetheless, effectively harnessing their potential for executing complex tasks crucially relies on enhancements in their function-calling capabilities. This paper identifies a critical gap in existing function-calling models, where performance varies significantly across benchmarks, often due to being misled by specific naming conventions. To address such an issue, we introduce Hammer, a novel family of foundation models specifically engineered for on-device function calling. Hammer employs an augmented dataset that enhances models' sensitivity to irrelevant functions and incorporates function masking techniques to minimize misleading. Our empirical evaluations reveal that Hammer not only outperforms larger models but also demonstrates robust generalization across diverse benchmarks, achieving state-of-the-art results. Our open-source contributions include a specialized dataset for irrelevance detection, a tuning framework for enhanced generalization, and the Hammer models, establishing a new standard for function-calling performance.
\end{abstract}

\section{Introduction}

Large language models (LLMs) have demonstrated remarkable proficiency in addressing a wide range of natural language processing tasks \citep{chowdhary2020natural}, as well as in handling long-context reasoning and complex planning \citep{wen2024reinforcing}. The use of LLMs as autonomous agents to assist humans in completing intricate tasks is increasingly in demand and is now more feasible from a technical standpoint than ever before \citep{gunter2024appleintelligencefoundationlanguage}. To fully capitalize on the potential of LLMs as autonomous agents, it is crucial for these models to accurately identify and utilize external tools or application programming interfaces (APIs), thereby enabling them to effectively execute complex tasks \citep{abdelaziz2024granite, patil2023gorilla}. Central to this capability is the model’s ability to select appropriate functions from a given set of options, provide accurate input arguments, and ultimately fulfill the user’s intent. Furthermore, in scenarios where no suitable function exists within the available options, the model must have the ability to decline the task, rather than making incorrect attempts \citep{patil2023gorilla}.

Recent advancements have introduced a variety of relevant datasets and benchmarks \citep{li2023apibankcomprehensivebenchmarktoolaugmented, wu2024sealtoolsselfinstructtoollearning}, along with the release of powerful models specifically designed for function-calling tasks \citep{zhang2024xlamfamilylargeaction, patil2023gorilla, abdelaziz2024granite}. Some models even simulate real-world scenarios, such as ticketing systems, to mimic more realistic use cases \citep{yao2024taubenchbenchmarktoolagentuserinteraction, chen2024travelagentaiassistantpersonalized}. Despite these significant strides in the development of function-calling models, our investigation reveals a critical gap: \textit{many existing models demonstrate considerable performance variations across different benchmarks}. As illustrated in \autoref{table:inconsistency}, this inconsistency underscores the need for further research into the robustness and generalization of function-calling models across diverse and practical task environments.

\begin{table}[h]
\scriptsize
\centering
\caption{Inconsistent performance of existing function-calling models across different benchmarks. For example, although xLAM-7B-fc achieved the best performance on most of the benchmarks, its performance significantly declined on the other two, resulting in the lowest average score overall.}
\label{table:inconsistency}
\begin{tabular}{c|ccccc|c}
\toprule
Models& BFCL & API-Bank & SealTool & Tool-Alpaca & Nexus Raven & Avg. \\ % Avg.
\midrule
Gorilla-OpenFunctions-v2-7B (FC) & \underline{79.1} & 62.5 & \underline{91.1} & 51.3 & \underline{68.4} & \underline{70.48}\\ % 70.48
Granite-20B-FunctionCalling (FC) & 76.63 & \underline{68.5} & \textbf{92.7} & \underline{58.0} & \textbf{75.1} & \textbf{74.186}\\ % 74.186
xLAM-7B-fc (FC) & \textbf{79.41} & \textbf{72.45} & 76.9 & \textbf{59.0} & 57.5 & 69.052\\ % 69.052
\bottomrule
\end{tabular}
\end{table}

Achieving such stability across diverse benchmarks is crucial, as it indicates the model's capability to generalize effectively to real-world applications \citep{yao2022webshop}. Driven by this objective, we begin by conducting a thorough analysis of the instability observed in existing models when executing function-calling tasks. Our findings highlight that one of the primary factors influencing generalization performance across benchmarks is the misleading nature of specific naming conventions for functions and parameters. Consequently, models tend to perform well on benchmarks that closely align with the naming conventions present in the training data but suffer notable performance declines when encountering benchmarks with differing naming styles. This problem will be examined in detail in Section~\ref{sec:existing-issues}.

In this paper, we present the \textbf{Hammer}, a family of lightweight models specifically fine-tuned for on-device function-calling tasks. This work is underpinned by a carefully designed irrelevance-augmented dataset and the use of function masking techniques, both aimed at enhancing the generalization capabilities of the models. To improve the models' ability to determine whether the user's intent aligns with the available function calls, we augment the xLAM-function-calling-60k dataset \citep{liu2024apigenautomatedpipelinegenerating} with an additional 7,500 instances specifically tailored for irrelevance detection. Furthermore, we introduce a function masking technique, which shifts the models' focus from function and parameter names to their descriptions, effectively reducing potential misinterpretations.

Following these advancements, Hammer demonstrates robust function-calling performance and strong generalization across a variety of benchmarks. Despite containing only 7 billion parameters, Hammer outperforms many larger open-source models and competes with top-tier closed-source models, such as GPT-4 \citep{achiam2023gpt} and GPT-4o \citep{islam2024gpt}, on the Berkeley Function Calling Leaderboard (BFCL) v2 \citep{berkeley-function-calling-leaderboard}. We benchmark Hammer and other models, including Salesforce's xLAM series \citep{zhang2024xlamfamilylargeaction} and IBM's Granite-20B-FunctionCalling \citep{abdelaziz2024granite}, across a range of representative datasets, such as API-Bank \citep{li2023apibankcomprehensivebenchmarktoolaugmented}, Tool-Alpaca \citep{tang2023toolalpaca}, Seal-Tools \citep{wu2024sealtoolsselfinstructtoollearning}, and Nexus Raven API Evaluation \citep{srinivasan2023nexusraven}. The results consistently highlight Hammer's exceptional generalization capabilities. The key contributions of our work could be summarized as follows:
\begin{itemize}
    \item \textbf{Tuning Framework:} A straightforward yet effective framework evolving function masking to tune function-calling models toward robust generalization capabilities, open-sourced at \url{https://github.com/MadeAgents/Hammer}.
    \item \textbf{Augmented Dataset:} A specialized dataset with 7,500 instances designed to enhance language models' awareness of irrelevance between candidate functions and user intent, open-sourced at \url{https://huggingface.co/datasets/MadeAgents/XLAM-7.5k-Irrelevance}.
    \item \textbf{Consistent SOTA Models:} Hammer, a family of well-trained function-calling models that demonstrate state-of-the-art performance across multiple benchmarks, open-sourced at \url{https://huggingface.co/MadeAgents/Hammer-7b}\footnote{Hammer-1.5B at \url{https://huggingface.co/MadeAgents/Hammer-1.5b}; Hammer-4B at \url{https://huggingface.co/MadeAgents/Hammer-4b}}.
\end{itemize}

\begin{table}[ht]
\scriptsize
\centering
\caption{Performance comparison of different models on Berkeley Function-Calling Leaderboard (as of date 09/20/2024). The rank is based on the overall accuracy, which is a weighted average of different evaluation categories. “FC" stands for function-calling mode in contrast to using a customized “Prompt" to extract the function calls. See ~\autoref{apx:extra-exp} for the complete list.}
\label{table:bfcl-summary}
\resizebox{\linewidth}{!}{
\begin{tabular}{ccccccc}
\toprule
Rank & Model & Overall Acc & AST Summary & Exec. Summary & Irrelevance & Relevance \\ 
\midrule
1 & GPT-4-0125-Preview (Prompt) & 85.79 & 85.50 & 89.25 & 61.35 & 97.56 \\
2 & GPT-4-1106-Preview (Prompt) & 85.00 & 86.31 & 87.38 & 64.98 & 90.24 \\
3 & GPT-4-0613 (Prompt) & 84.74 & 84.66 & 87.57 & 75.57 & 82.93 \\
\rowcolor[HTML]{EFEFEF} & Hammer-7B (FC) & 83.92 & 78.70 & 89.72 & 72.87 & 92.68 \\
4 & GPT-4-turbo-2024-04-09 (Prompt) & 83.89 & 85.41 & 88.13 & 61.82 & 82.93 \\
5 & GPT-4o-mini-2024-07-18 (Prompt) & 83.35 & 80.52 & 87.95 & 79.20 & 80.49 \\
7 & Functionary-Medium-v3.1-70B (FC) & 82.55 & 81.06 & 89.32 & 73.23 & 70.73 \\
13 & Functionary-Small-v3.1-8B (FC) & 80.21 & 78.64 & 83.45 & 68.36 & 85.37 \\
16 & xLAM-7B-fc (FC) & 79.41 & 72.77 & 85.68 & 79.76 & 80.49 \\
19 & Gorilla-OpenFunctions-v2-7B (FC) & 79.10 & 73.18       & 84.97 & 73.13 & 85.37 \\
21 & Functionary-Small-v3.2-8B (FC) & 78.96 & 76.16 & 83.04 & 72.32 & 80.49 \\
25 & FireFunction-v2-70B (FC) & 77.45 & 74.20 & 84.23 & 52.94 & 87.80 \\
26 & Granite-20B-FunctionCalling (FC) & 76.63 & 66.73       & 82.97 & 72.43 & 95.12 \\
\rowcolor[HTML]{EFEFEF} & Hammer-4B (FC) & 76.05 & 69.59 & 80.82 & 68.66 & 90.24 \\
31 & xLAM-1.3B-fc (FC) & 74.90 & 67.37 & 80.80 & 61.21 & 95.12 \\
32 & Hermes-2-Pro-Llama-3-70B (FC) & 74.78 & 72.09 & 81.29 & 53.80 & 80.49 \\
\rowcolor[HTML]{EFEFEF} & Hammer-1.5B (FC) & 73.04 & 65.53 & 75.86 & 72.18 & 92.68 \\
40 & Command-R-Plus (FC) & 72.04 & 66.32 & 77.41 & 52.75 & 92.68 \\
45 & Hermes-2-Pro-Llama-3-8B (FC) & 66.18 & 64.18 & 74.05 & 55.16 & 53.66 \\
46 & Hermes-2-Pro-Mistral-7B (FC) & 65.44 & 60.82 & 74.25 & 38.55 & 75.61 \\
47 & Hermes-2-Theta-Llama-3-8B (FC) & 64.83 & 61.08 & 72.54 & 62.66 & 51.22 \\
57 & FireFunction-v1-46B (FC) & 48.11 & 38.16 & 41.20 & 68.55 & 95.12 \\
% 68 & Hermes-2-Theta-Llama-3-70B (FC) & 10.00 & 0.00 & 0.00 & 100.00 & 0.00 \\ 
\bottomrule
\end{tabular}
}
\end{table}

\begin{table}[ht]
\scriptsize
\centering
\caption{Performance comparison of different models on several academic benchmarks. The rank is based on the average F1 score on ``Func. + Args'', which indicates both function selection and parameter filling are accurate.}
\label{table:ibm-summary}
\resizebox{\linewidth}{!}{
\begin{tabular}{cccccc|cc}
\toprule
& \multicolumn{5}{c|}{F1 Func-Name $|$ F1 Func. + Args} & \multicolumn{2}{c}{F1 Average} \\ 
\midrule
% \cline{3-14} \\
\multirow{-1}{*}{Model} & \begin{tabular}[c]{@{}c@{}}API-Bank\\ L-1\end{tabular} & \begin{tabular}[c]{@{}c@{}}API-Bank\\ L-2\end{tabular} & Tool-Alpaca & \begin{tabular}[c]{@{}c@{}}Seal-Tools\\ (Single-Tool)\end{tabular} & \begin{tabular}[c]{@{}c@{}}Nexus\\ Raven\end{tabular} & \begin{tabular}[c]{@{}c@{}}Func\\ Name\end{tabular} & \begin{tabular}[c]{@{}c@{}}Func.+\\ Args\end{tabular} \\ 
\midrule
 GPT-4-0613 (Prompt) &92.93 $|$ 84.78 & 69.60 $|$ 56.98 & 88.64 $|$ 66.67 & 94.56 $|$ 93.95 & 95.73 $|$ 91.60 & 88.29 & 78.79 \\
GPT-4o-mini (Prompt) & 95.08 $|$ 89.28 & 84.35 $|$ 67.52 & 64.34 $|$ 54.69 & 87.94 $|$ 86.00 & 91.72 $|$ 84.59 & 84.69 & 76.42 \\
\rowcolor[HTML]{EFEFEF} Hammer-7B (FC) & 93.48 $|$ 85.79 & 82.91 $|$ 66.40 & 82.31 $|$ 59.86 & 97.44 $|$ 91.66 & 92.46 $|$ 77.35 & 89.72 & 76.21 \\
Granite-20B-FunctionCalling (FC) & 90.41 $|$ 77.82 & 78.95 $|$ 59.15 & 77.27 $|$ 58.00 & 94.86 $|$ 92.70 & 94.47 $|$ 75.14 & 87.19 & 72.56 \\
\rowcolor[HTML]{EFEFEF} Hammer-4B (FC) & 91.65 $|$ 81.46 & 77.59 $|$ 61.01 & 85.09 $|$ 56.96 & 96.42 $|$ 92.45 & 81.73 $|$ 64.89 & 86.50 & 71.35 \\
xLAM-7B-fc (FC) & 90.05 $|$ 80.69 & 72.49 $|$ 64.24 & 67.26 $|$ 58.96 & 78.97 $|$ 76.87 & 54.09 $|$ 57.50 & 72.57 & 67.65 \\
Gorilla-OpenFunctions-v2-7B (FC) & 69.21 $|$ 70.34 & 48.82 $|$ 54.69 & 72.93 $|$ 51.26 & 93.20 $|$ 91.11 & 72.84 $|$ 68.41 & 71.40 & 67.16 \\
xLAM-1.3B-fc (FC) & 94.86 $|$ 83.70 & 91.80 $|$ 64.32 & 64.86 $|$ 50.58 & 90.74 $|$ 80.43 & 64.43 $|$ 54.80 & 81.34 & 66.77 \\
\rowcolor[HTML]{EFEFEF} Hammer-1.5B (FC) & 82.13 $|$ 72.30 & 79.82 $|$ 59.71 & 80.90 $|$ 53.48 & 95.59 $|$ 88.65 & 79.87 $|$ 56.88 & 83.66 & 66.20 \\ 
Qwen2-7B-Instruct (Prompt) & 81.55 $|$ 60.62 & 95.65 $|$ 49.50 & 71.59 $|$ 48.11 & 93.88 $|$ 77.51 & 87.05 $|$ 63.47 & 85.94 & 59.84 \\
Qwen2-1.5B-Instruct (Prompt) & 74.63 $|$ 63.55 & 57.69 $|$ 33.62 & 65.76 $|$ 45.25 & 82.08 $|$ 75.49 & 70.62 $|$ 45.46 & 70.16 & 52.67 \\
Qwen1.5-4B-Chat (Prompt) & 55.33 $|$ 59.78 & 46.74 $|$ 38.48 & 35.41 $|$ 16.98 & 48.44 $|$ 62.32 & 29.03 $|$ 33.70 & 42.99 & 42.25 \\
\bottomrule
\end{tabular}}
\end{table}

\section{Related Works}

% \jwliao{1. LLMs for sequential task solving as an agent; 2. LLMs dedicated to function calling or tool using, datasets and benchmarks; 3. current optimization methods and their common feature (SFT alike?). 4. Interactive, multi-turn, nested fc and alike multi-dimensional research.}

\textbf{LLMs as Agents for Function-Calling.}
Recent research has shown significant interest in leveraging LLMs as autonomous agents to perform complex tasks through function calling and tool usage \citep{erdogan2024tinyagent, chen2024octopus}. IBM Granite-20B-FunctionCalling model \citep{abdelaziz2024granite} proposes a multi-task learning framework trained on seven core function-calling tasks, demonstrating superior performance over other open models on the BFCL v2 benchmark. APIGen adopts an automated pipeline for generating high-quality, diverse function-calling datasets, building a 7B-parameter model to surpass GPT-4’s performance \citep{liu2024apigenautomatedpipelinegenerating}. Similarly, ToolACE \citep{liu2024toolacewinningpointsllm} generates diverse tool-learning datasets, allowing its 8B-parameter ToolACE-8B model to achieve state-of-the-art results on the BFCL v2, rivaling the latest GPT-4 models. Further studies explore various dimensions of function calling, such as improving efficiency through parallel function calls \citep{amechanismoffunctioncallsinMSVL}, identifying vulnerabilities in function calling processes \citep{srinivasan2023nexusraven}, and developing benchmarks to evaluate LLMs' ability to handle diverse function calls \citep{pmlr-v235-kim24y}. Collectively, this body of work emphasizes the role of function calling in enabling LLMs to act autonomously and integrate external tools and resources effectively.

\textbf{Datasets and Benchmarks for Function-Calling Evaluation.}
Substantial advancements have been made in developing datasets and benchmarks to assess the function-calling capabilities of LLMs. API-BLEND \citep{basu2024apiblendcomprehensivecorporatraining} introduces a large corpus for training and systematically testing tool-augmented LLMs. It includes real-world scenarios involving API-related tasks such as API/tool detection, slot filling, and sequencing of detected APIs. API-Bank \citep{li2023apibankcomprehensivebenchmarktoolaugmented} provides a comprehensive dataset featuring 2,138 distinct APIs and 1,888 dialogues with 4,149 API calls. This dataset is designed to evaluate LLMs' tool-utilization capabilities, including planning, retrieval, and API-calling proficiency. APIGen \citep{liu2024apigenautomatedpipelinegenerating} employs an automated and rigorous data generation process to create a diverse dataset that includes various query styles, such as parallel function calling, and undergoes a multi-stage verification process to ensure data accuracy and relevance. Seal-Tools \citep{wu2024sealtoolsselfinstructtoollearning} introduces a large-scale, self-instruct API-like tool-learning dataset that incorporates practical application scenarios and nested tool calls.

\textbf{Tuning Techniques for Function-Calling Models.}
IBM’s Granite-20B-FunctionCalling model is trained using a multi-task learning approach, which enables language models to develop function-calling capabilities by mastering a range of granular tasks \citep{abdelaziz2024granite}. TinyAgent focuses on equipping small language models (SLMs) with complex reasoning and function-calling abilities, allowing for secure and private deployment at the edge. It employs LoRA fine-tuning, incorporates negative samples, and uses in-context examples selected via retrieval-augmented generation (RAG) \citep{gao2023retrieval} to enhance function selection and orchestration accuracy through directed acyclic graph (DAG) comparison \citep{tiny-agent}. The xLAM series utilizes a supervised fine-tuning (SFT) approach with direct preference optimization (DPO) \citep{rafailov2024direct} alignment, integrating data parallelism, LoRA, and a cosine learning rate scheduler to optimize performance across various categories of function-calling agents \citep{zhang2024xlamfamilylargeaction}.

\section{Problem Statement and Analysis}
\label{sec:existing-issues}

This section aims to introduce and analyze the common challenges that function-calling models encounter in practical applications. Through this analysis, we seek to identify methods to enhance models' stability and generalization capabilities in real-world scenarios. 

% 更加compact一点，体现出多个candidates中做选择的过程
% 统一叫args，文中说明其根param的关系。
\begin{figure*}[ht]
\begin{center}
\centerline{\includegraphics[width=.95\columnwidth]{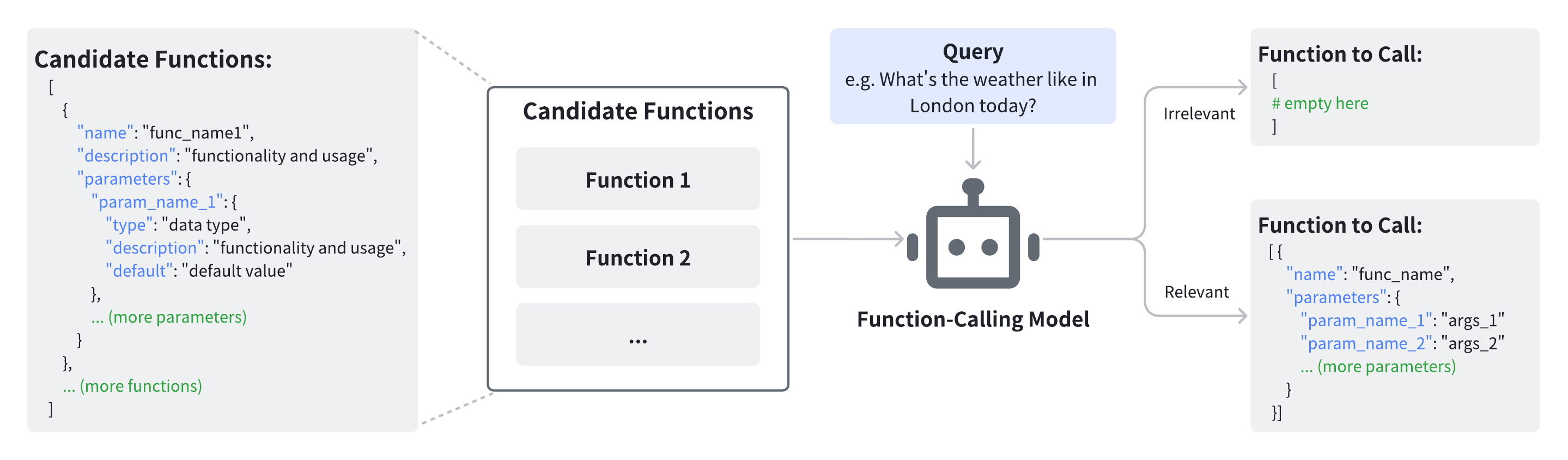}}
\caption{\normalsize Demonstration of a simple function-calling process.}
\label{fig:function-call-demo}
% \vspace{-2.5em}
\end{center}
\end{figure*}

Before delving into the specific issues, we present a typical function-calling process, illustrated in \autoref{fig:function-call-demo}. In this process, each candidate encompasses several components, including the function name, parameter names, default values, and descriptions. The objective of the model is to output complete and accurate function-calling code that can accomplish users' intent or, alternatively, output an empty list to indicate that none of the given candidates can satisfy the user's requirements \citep{berkeley-function-calling-leaderboard}. Achieving this goal hinges on the model's ability to accurately align the user's intent with the functionality of the candidate functions, i.e., selecting the appropriate function, and its capacity to comprehend the usage of each parameter, i.e., populating the function with the correct arguments. However, certain recurring issues have been observed in practice.

\subsection{Misleadingness by Function Name and Parameter Name}
\label{sec:misleading-issue}

As illustrated in \autoref{fig:function-call-demo}, the definition of a function typically comprises the function name, parameter names, and descriptions. The format of function and parameter names is often quite compact, e.g., cal\_sum or max\_value, and influenced by the designer's personal style and preferences. When a model attempts to infer the function's purpose solely from the function name, this compactness can lead to ambiguities, misguiding the model's selection, particularly in the presence of complex functionalities \citep{gunter1992semantics}. For instance, a function named \texttt{parse\_data} might be intended for parsing JSON data, but the same name could refer to parsing CSV files in a different context, leading to potential misinterpretations. Similarly, when deducing the usage of parameters based on their names, models may be misled by the historical usage of similarly named parameters in the training dataset. More specifically, these misleading scenarios can be categorized into several cases.

\textbf{Misled by Function Names.} When a user intent aligns closely with a function name present in the training labels, the model may incorrectly prioritize that function from the candidate list during testing, even if its functionality diverges significantly from the intended operation. For example, if a function named \texttt{fetch\_data} is included in the training pairs for retrieving user data from a database, but in the testing set, a function with the same name retrieves data from an external API, the model may erroneously select it based solely on the name.

\textbf{Misled by Parameter Names.} In instances where the functionality and descriptions of parameters change within the testing environment, the model frequently clings to its original patterns of parameter usage, resulting in incorrect function calls. For instance, if a function's parameter \texttt{timeout} is expected to be an integer representing seconds in one context, but in another context, it is defined as a string in the format ``10s", the model's reliance on the original integer format may lead to erroneous calls.

\textbf{Disturbed by Naming Preferences.} The model's robustness can diminish when the naming conventions of functions or parameters in the testing environment diverge from those in the training dataset. Variations, such as discrepancies between \texttt{CamelCase} and \texttt{snake\_case} may adversely lower the model's confidence, as an on-device lightweight model may struggle to generalize across different naming styles.

% In contrast, the descriptions section provides a more flexible natural language explanation, often encapsulating the information that function and parameter names aim to convey. While descriptions can also reflect the designer’s personal style to some extent, they tend to be more accurate and detailed, thus reducing the likelihood of ambiguity or misguidance. Consequently, when training function-calling models, we face the challenge of not knowing the preferences or naming styles of function designers in real-world applications. Thus, it is reasonable to expect that the trained model should minimize its reliance on naming conventions of function names and parameter names, as these are highly susceptible to individual variations. That is, the model should rely on understanding the function’s purpose and usage through its description rather than attempting to infer functionality based on potentially ambiguous, compact components such as function and parameter names.

% By focusing on the description, the model can more accurately grasp the function’s intent and expected behavior, ensuring robust performance across diverse naming conventions and avoiding pitfalls introduced by overfitting to specific naming patterns in the training data. It also promotes better generalization, as the description typically offers a more comprehensive view of the function's role, beyond what can be conveyed by concise names.

\subsection{The Impact of Excessive Focus on the Naming}

\begin{figure*}[ht]
    \centering
    \begin{subfigure}{0.48\linewidth}
        \centering
        \includegraphics[width=2.9in]{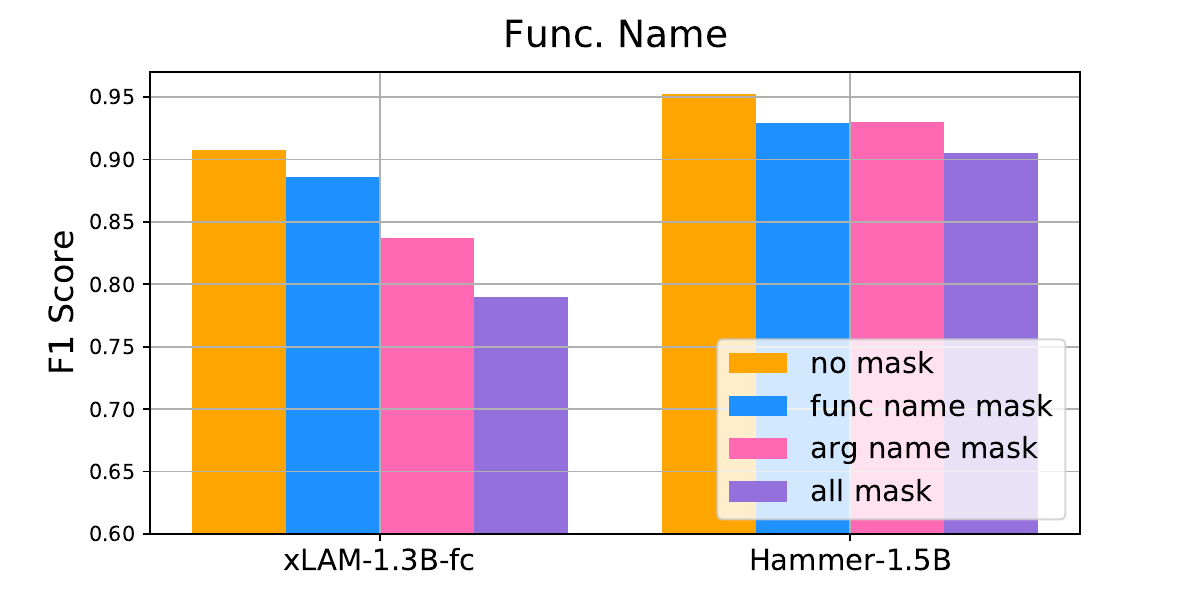}
    \end{subfigure}
    % \hspace{3em}
    \begin{subfigure}{0.48\linewidth}
        \centering
        \includegraphics[width=2.9in]{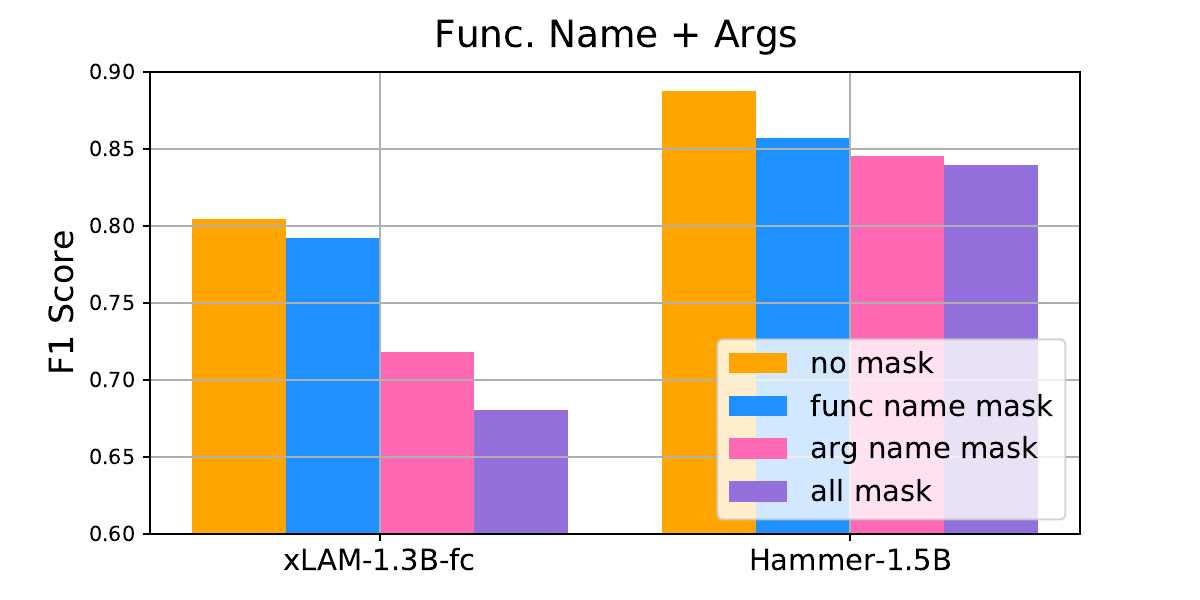}
    \end{subfigure}
    \caption{\normalsize Case studies examining the performance degradation when function names and parameter names are obfuscated during test time.} 
    \label{fig:issue-overfit}
\end{figure*}

To investigate the extent to which existing models rely on function and parameter names and corresponding impact, we conducted a case study using the xLAM-1B-fc model on the Seal-Tools benchmark. Specifically, we masked the function and parameter names in the test set, i.e., replaced them with random strings, and observed how the model's performance changed. As shown in \autoref{fig:issue-overfit}, after masking the function and parameter names, even though the descriptions contained all necessary information about the function's purpose and usage, the performance of xLAM-1B-fc dropped significantly. This result confirms the model's overreliance on function and parameter names, highlighting the potential risks this behavior may pose in real-world applications.

In contrast, \autoref{fig:issue-overfit} also presents the performance of our Hammer model under the same setting. Hammer exhibited a much smaller performance drop, demonstrating its robustness when faced with arbitrary function and parameter naming patterns. This resilience suggests that Hammer relies more heavily on the function descriptions rather than compact, potentially ambiguous names. In Section~\ref{sec:method}, we will provide a detailed explanation of Hammer's training methodology.

\section{Methodology}
\label{sec:method}

In this section, we describe our detailed approach and augmented dataset to fine-tune the Hammers, a series of robust language models designed for function-calling. 

\subsection{Function Masking}
% \muning{A figure of the entire training framework is here. highlight the balance of irrelevance and function calling. distinguish the relationship between irrelevance and function calling}

In light of the analysis in Section~\ref{sec:existing-issues}, a direct approach to mitigate these issues involves minimizing the interference from function names and parameter names, while enforcing the model to comprehend the functionality and usage of candidates based on their descriptions. In contrast, the descriptions provide a more flexible natural language explanation, often encapsulating the information that function and parameter names aim to convey. While descriptions can also reflect the designer’s personal style to some extent, they tend to be more accurate and detailed, thus reducing the likelihood of ambiguity or misguidance. Consequently, when training function-calling models, we face the challenge of not knowing the preferences or naming styles of function designers in real-world applications. Thus, it is reasonable to expect that the trained model should understand the function’s purpose and usage through its description rather than attempting to infer functionality based on potentially ambiguous, compact components such as function and parameter names.

\begin{figure*}[ht]
\begin{center}
\centerline{\includegraphics[width=.95\columnwidth]{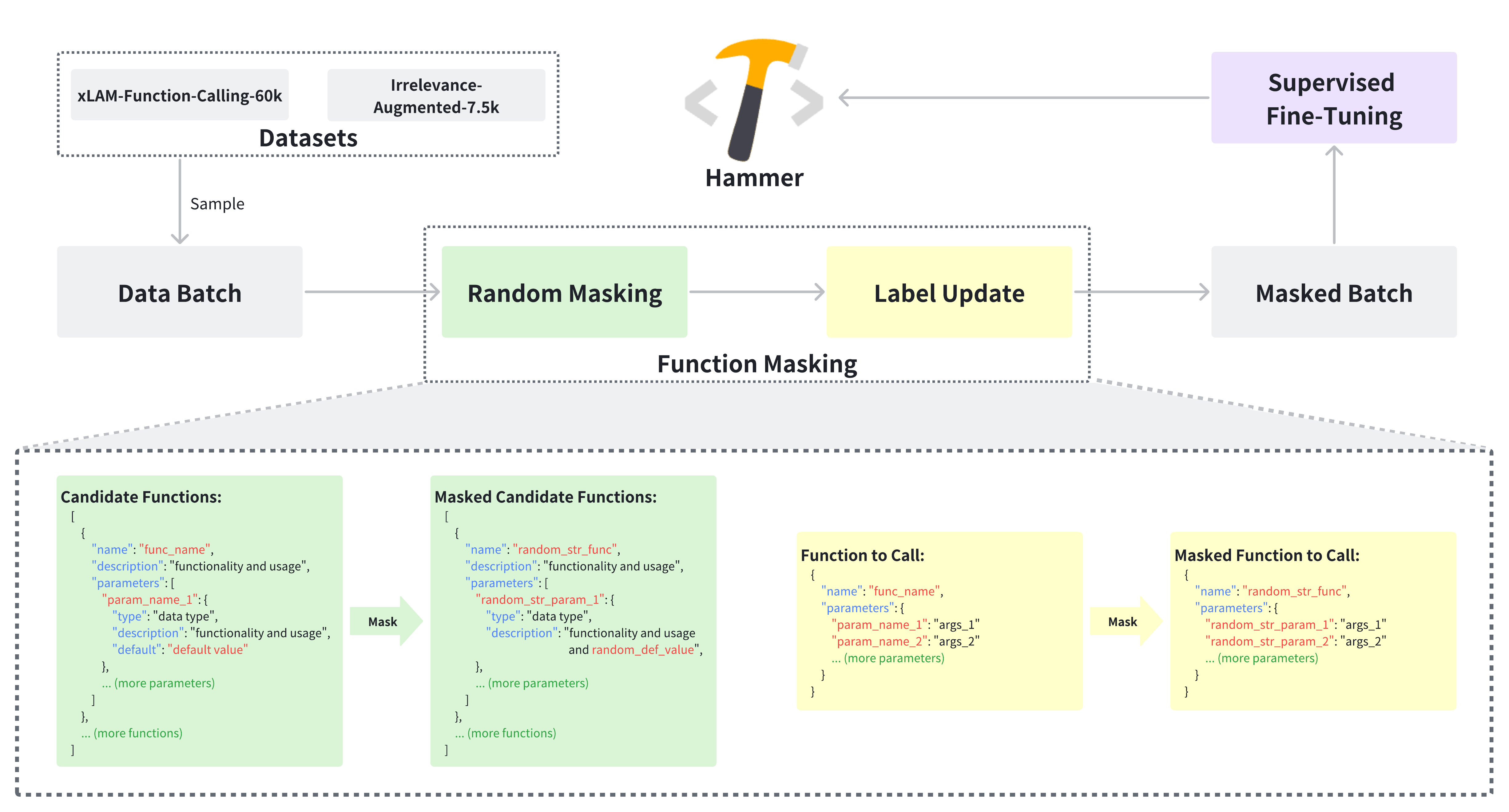}}
\caption{\normalsize Step-by-step building workflow of Hammer series with function masking.}
\label{fig:function-masking-process}
% \vspace{-2.5em}
\end{center}
\end{figure*}

To this end, we propose a tuning framework for function-calling models based on a masking mechanism, with a full pipeline shown in \autoref{fig:function-masking-process}. This framework aims to guide the model's attention toward the description, and thus enhance the model's generalization capabilities in practice. Specifically, in our proposed framework:
\begin{itemize}
    \item \textbf{Function names} in candidates are masked by replacing them with randomly generated strings during training. This technique minimizes the model’s reliance on memorizing function names, prompting it to understand the function’s purpose solely through its description. By doing so, the model becomes more adaptable across various coding practices, as it is less influenced by common naming conventions.
    \item \textbf{Parameter names} in candidates are substituted with random strings as well, ensuring the model focuses on the parameter descriptions rather than the specific names, which often vary between implementations.
    \item \textbf{Default parameter values} in candidates are randomized and appended to the parameter descriptions. This also guides models to pay more attention to the parameter descriptions.
    \item \textbf{Labels} in batch are updated according to the masked candidate list, i.e. replacing the function name and parameter name with corresponding masked strings in candidates.
\end{itemize}

By focusing on the description, the model can more accurately grasp the function’s intent and expected behavior, ensuring robust performance across diverse naming conventions and avoiding pitfalls introduced by overfitting to specific naming patterns in the training data. It also promotes better generalization, as the description typically offers a more comprehensive view of the function's role, beyond what can be conveyed by concise names.

% Through these mechanisms, we aim to bolster the model's robustness and enable it to generalize better across various scenarios, ultimately improving its effectiveness in real-world function-calling tasks.

\subsection{Irrelevance-Augmented Dataset}

During the fine-tuning process using the xlam-function-calling-60k dataset, we identified a concerning inverse relationship between the model's ability to accurately execute function calls and its capacity for irrelevance detection—specifically, the ability to assess whether there exists no function call in the candidate set aligns with the user’s intent. The details of this observation are discussed further in Section~\ref{sec:ablation-irrelevance}. This phenomenon indicates that, while fine-tuning lightweight language models on datasets specialized in function selection can improve their accuracy in choosing appropriate functions from a predefined set, it may unintentionally impair their ability to detect irrelevance. As a result, models might generate inappropriate function calls, even in the absence of valid options.

To address this issue, we propose an irrelevance-augmented dataset. This augmentation, applied to the original xlam-function-calling-60k dataset, incorporates 7,500 examples sampled from the original training set, in which the correct functions are deliberately excluded from the candidate list, and the labels are replaced with empty lists. By exposing the model to more instances requiring irrelevance detection, we aim to enhance its ability to discern when to abstain from making function calls, thus promoting more judicious function selection.

% \ngy{Experiments: qwen2 1.5 B (Train bs=16, lr=5e-5, lora rank=32) 10000 data in total, different rejection ratio. E.g., 5\% means 500 rejection data, 9500 non-reject data. How these 9500 data generated: first random sample 10000 non-reject data from 60k, then random sample 9500. This can reduce variance since we use similar non-reject data for all plotted points. ast\_ratio.png indicates a negative relation between fc ability and ratio. Overall\_ratio indicates there is an optimal ratio balancing fc and rejection. The curve for Overall\_ratio is fitted using 2 piecewise quadratic curves. Emphasize this phenomenon only happens for small models.}

\section{Evaluation}

In this section, we show the superiority of our Hammers in performance and robustness across various benchmarks, and in-depth analysis to verify the effectiveness of our augmented dataset and approach.

% In this section, we show the superiority of our Hammers in performance and robustness. After briefly introducing involved benchmarks and metrics, we present Hammers' overall performance across these benchmarks and subsequently delve deeper to investigate the varying levels of their distinct capabilities. Furthermore, we conduct a series of meaningful ablations: applying our training methodology and augmented datasets to different baseline models to validate that they are not only applicable to the Qwen2 series; examining the impact of varying ratios of irrelevant data to original data on model performance in an effort to identify an optimal ratio; and analyzing the effects of different masking ratios on the model's generalization ability. 
% \muning{More training details can be found in the Appendix.}

\subsection{Experimental Setup}

\textbf{Benchmarks.} To assess the generalizability of Hammers, we conducted evaluations using a variety of function-calling benchmarks, all of which represent out-of-domain challenges for our model. The \textbf{Berkeley Function-Calling Leaderboard (BFCL)} \citep{berkeley-function-calling-leaderboard} provides a comprehensive dataset comprising over 1,700 instances. It covers tasks such as Simple Function, Multiple Function, Parallel Function, and Parallel Multiple Function for Python, as well as function relevance detection, REST API, JavaScript, and Java for non-Python environments. \textbf{API-Bank} \citep{li2023apibankcomprehensivebenchmarktoolaugmented}, consisting of 314 tool-use dialogues and 753 API calls, evaluates models' ability to correctly invoke a known API (L-1) based on a query, and to retrieve and call APIs from a candidate list (L-2). Similarly, \textbf{Nexus Raven API Evaluation} \citep{srinivasan2023nexusraven} offers 318 test examples across 65 distinct APIs, contributing further to the evaluation of function-calling capabilities. \textbf{Tool-Alpaca} \citep{tang2023toolalpaca} employs a synthetic data generation method, featuring 271 tool-use instances in 50 categories. For evaluation, we utilized 100 simulated test examples from this dataset, similar to Nexus Raven. Lastly, \textbf{Seal-Tools} \citep{wu2024sealtoolsselfinstructtoollearning} represents one of the most extensive and recent benchmarks, with 4,076 automatically generated APIs across various life domains. As one of the newest benchmarks, Seal-Tools presents a lower risk of data leakage. 

\textbf{Evaluation Metrics.} BFCL assesses function-calling models through two primary evaluation methods: Abstract Syntax Tree (AST) Evaluation and Executable Function Evaluation \citep{berkeley-function-calling-leaderboard}. The AST evaluation emphasizes the syntactic precision of the generated function calls, ensuring that the model's output adheres to a predefined function documentation in terms of structure and parameters. This includes verifying the correctness of function names, required parameters, and appropriate data types. In contrast, Executable Function Evaluation takes this further by executing the generated function calls to assess their functional accuracy. This evaluation ensures that the functions not only compile but also run correctly, producing the intended outputs, which is vital for real-world applications. In addition to BFCL, we incorporated F1 scores to measure exact matches of API names and parameters to evaluate the models on alternative benchmarks \citep{abdelaziz2024granite}. 

\subsection{Overall Performance on Various Benchmarks}

We first evaluate Hammer series on BFCL. \autoref{table:bfcl-summary} indicates that within the BFCL framework, our Hammer series consistently achieves corresponding sota performance at comparable scales, particularly Hammer-7B, whose overall performance ranks second only to the proprietary GPT-4. In addition, we evaluated our Hammer series (1.5b, 4b, 7b) on other academic benchmarks to further show our model's generalization ability. Upon observing Hammer's performance across various benchmarks unrelated to the xlam-function-calling-60k Datasets, as shown in \autoref{table:ibm-summary}, we find that Hammer demonstrates remarkably stable performance, which indicates the robustness of Hammers.

\subsection{Detailed Performance on Different Types of Function Calling}

\begin{figure*}[ht]
\centering
\centerline{\includegraphics[width=.99\columnwidth]{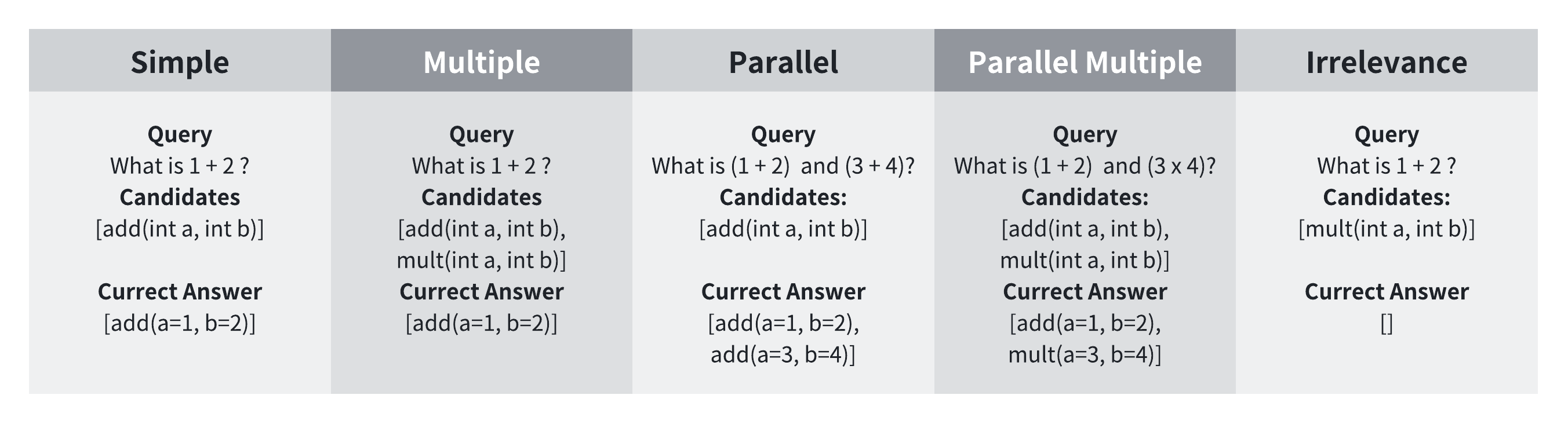}}
\caption{\normalsize Demonstration of different function-calling tasks.}
\label{fig:function-calling-style}
\end{figure*}

In this section, we closely examine the performance of Hammer across different types of function-calling tasks, as exampled in \autoref{fig:function-calling-style}, and detailed in \autoref{apx:func-type}.

As shown in \autoref{table:bfcl-ast-exec}, we found that Hammer-7B demonstrates exceptional overall performance across these various tasks. Its AST Summary is second only to the GPT-4 series and Functionary-Medium-v3.1-70B. Notably, Hammer-7B even outperformed GPT-4 in the more practically relevant Executable Function Evaluation, highlighting the potential of Hammer and its function-masking training strategy in real-world scenarios. Moreover, we observed that Hammer-7B achieved state-of-the-art results in both AST Evaluation and Executable Function Evaluation for the most complex Parallel Multiple task. This suggests that the function-masking training approach becomes increasingly advantageous as task complexity rises. This aligns with our insight that more complex tasks typically demand a deeper understanding of functions, necessitating models to focus more on function descriptions.

\begin{table}[ht]
\scriptsize
\centering
\caption{Detailed performance comparison of different models using AST evaluation and Executable Function Evaluation with regard to four function-calling styles on BFCL (as of date 09/20/2024).}
\label{table:bfcl-ast-exec}
\begin{tabular}{c|ccccc|c}
\toprule
& \multicolumn{5}{|c|}{AST+Exec. (AST Acc. $|$ Exec. Acc.)} & Acc. \\
\midrule
Model & Summary & Simple & Multiple & Parallel & Parallel Multiple & Irrelevance \\ 
\midrule
GPT-4-0125-Preview (Prompt) & 85.50 $|$ 89.25 & 78.82 $|$ \textbf{99.00} & 88.44 $|$ \textbf{96.00} & 91.00 $|$ 82.00 & 83.75 $|$ 80.00 & 61.35\\
GPT-4-1106-Preview (Prompt) & \textbf{86.31} $|$ 87.38 & 78.75 $|$ \textbf{99.00} & \textbf{89.12} $|$ \textbf{96.00} & \textbf{94.12} $|$ 82.00 & 83.25 $|$ 72.50 & 64.98\\
GPT-4-0613 (Prompt) & 84.66 $|$ 87.57 & 78.76 $|$ 98.29 & 85.46 $|$ \textbf{96.00} & 91.75 $|$ 86.00 & 82.67 $|$ 70.00 & 75.57\\
\rowcolor[HTML]{EFEFEF} Hammer-7B (FC) & 78.70 $|$ \textbf{89.72} & 69.31 $|$ 91.86  & 82.52 $|$ 94.00 & 78.88 $|$ 88.00 & \textbf{84.08} $|$ \textbf{85.00} & 72.87\\
GPT-4-turbo-2024-04-09 (Prompt) & 85.41 $|$ 88.13 & \textbf{80.47} $|$ \textbf{99.00} & 88.81 $|$ \textbf{96.00} & 88.12 $|$ 80.00 & 84.25 $|$ 77.50 & 61.82\\
GPT-4o-mini-2024-07-18 (Prompt) & 80.52 $|$ 87.95 & 75.88 $|$ 98.29 & 81.64 $|$ 94.00 & 85.12 $|$ 82.00 & 79.42 $|$ 77.50 & 79.20\\
Functionary-Medium-v3.1-70B (FC) & 81.06 $|$ 89.32 & 74.34 $|$ 98.29 & 87.59 $|$ 94.00 & 81.62 $|$ \textbf{90.00} & 80.67 $|$ 75.00 & 73.23\\
Functionary-Small-v3.1-8B (FC) & 78.64 $|$ 83.45 & 72.70 $|$ 87.79 & 83.31 $|$ 90.00 & 85.62 $|$ 86.00 & 72.92 $|$ 70.00 & 68.36\\
xLAM-7B-fc (FC) & 72.77 $|$ 85.68 & 70.28 $|$ 94.21 & 78.18 $|$ 88.00 & 74.12 $|$ 88.00 & 68.50 $|$ 72.50 & \textbf{79.76}\\
Gorilla-OpenFunctions-v2-7B (FC) & 73.18 $|$ 84.97 & 70.81 $|$ 95.86 & 79.47 $|$ \textbf{96.00} & 75.75 $|$ 78.00 & 66.67 $|$ 70.00 & 73.13\\
Functionary-Small-v3.2-8B (FC) & 76.16 $|$ 83.04  & 69.50 $|$ 90.64 & 81.50 $|$ 88.00 & 80.12 $|$ 86.00 & 73.50 $|$ 67.50 & 72.32\\
FireFunction-v2-70B (FC) & 74.20 $|$ 84.23 & 74.11 $|$ 94.43 & 81.49 $|$ 88.00 & 73.62 $|$ 82.00 & 67.58 $|$ 72.50 & 52.94\\
Granite-20B-FunctionCalling (FC) & 66.73 $|$ 82.97 & 65.27 $|$ 85.36 & 73.05 $|$ 90.00 & 60.75 $|$ 84.00 & 67.83 $|$ 72.50 & 72.43\\
\rowcolor[HTML]{EFEFEF} Hammer-4B (FC) & 69.59 $|$ 80.82 & 62.58 $|$ 67.79 & 77.72 $|$ 92.00 & 69.12 $|$ 86.00 & 68.92 $|$ 77.50 & 68.66\\
xLAM-1.3B-fc (FC) & 67.37 $|$ 80.80 & 64.49 $|$ 79.21 & 73.06 $|$ 88.00 & 64.00 $|$ 86.00 & 67.92 $|$ 70.00 & 61.21\\
Hermes-2-Pro-Llama-3-70B (FC) & 72.09 $|$ 81.29 & 66.29 $|$ 80.64 & 73.49 $|$ 88.00 & 70.25 $|$ 84.00 & 78.33 $|$ 72.50 & 53.80\\
\rowcolor[HTML]{EFEFEF} Hammer-1.5B (FC) & 65.53 $|$ 75.86 & 62.34 $|$ 49.93 & 72.84 $|$ 92.00 & 58.75 $|$ 84.00 & 68.17 $|$ 77.50 & 72.18\\
Command-R-Plus (FC) & 66.32 $|$ 77.41 & 64.25 $|$ 89.14 & 72.45 $|$ 86.00 & 66.25 $|$ 82.00 & 62.33 $|$ 52.50 & 52.75\\
Hermes-2-Pro-Llama-3-8B (FC) & 64.18 $|$ 74.05 & 62.32 $|$ 68.71 & 74.96 $|$ 90.00 & 61.62 $|$ 80.00 & 57.83 $|$ 57.50 & 55.16\\
Hermes-2-Pro-Mistral-7B (FC) & 60.82 $|$ 74.25 & 60.98 $|$ 60.50 & 71.49 $|$ 90.00 & 60.38 $|$ 84.00 & 50.42 $|$ 62.50 & 38.55\\
Hermes-2-Theta-Llama-3-8B (FC) & 61.08 $|$ 72.54 & 58.53 $|$ 69.14 & 67.82 $|$ 88.00 & 59.62 $|$ 78.00 & 58.33 $|$ 55.00 & 62.66\\
\bottomrule
\end{tabular}
\end{table}

\subsection{Ablation on Different Base Models} 

To further validate the effectiveness of our augmented data and tuning technique, we applied our approach to two different sizes of the deepseek-coder models, in addition to the Qwen series. The results are illustrated in \autoref{table:different-base-model}. Upon examining the results presented in the table, we first note that the fine-tuned Hammer model exhibits a notable performance improvement compared to the vanilla Qwen model \citep{bai2023qwen, yang2024qwen2}, thereby confirming the efficacy of our data and methodology on the Qwen architecture. Subsequently, we compared the deepseek-coder model \citep{guo2024deepseek} before and after fine-tuning; the fine-tuned variant, referred to as deepseek-coder-Hammer, demonstrates significant enhancements over the vanilla model, despite the poor performance of deepseek-coder prior to fine-tuning. This suggests that our methodology is not exclusively applicable to the Qwen model. Furthermore, it is noteworthy that the performance of the deepseek-coder-Hammer, fine-tuned using our approach, significantly surpasses that of the xLAM model, which was also based on deepseek-coder-instruct and obtained through SFT with the xlam-function-calling-60k dataset. This further underscores the superiority of our proposed method.

\begin{table}[ht]
\scriptsize
\centering
\caption{\normalsize Ablation on different base models and benchmarks. Except Qwen series, We also apply the same tuning process to Deepseek-Coder-1.3B-Instruct and Deepseek-Coder-6.7B.}
\label{table:different-base-model}
\begin{tabular}{cccccc|cc}
\toprule
& \multicolumn{5}{c}{Academic Benchmarks (F1 Func-Name $|$ F1 Func. + Args)} & \multicolumn{2}{c}{F1 Average} \\
\midrule
\multirow{-1}{*}{Model} & \begin{tabular}[c]{@{}c@{}}API-Bank\\ L-1\end{tabular} & \begin{tabular}[c]{@{}c@{}}API-Bank\\ L-2\end{tabular} & Tool-Alpaca & \begin{tabular}[c]{@{}c@{}}Seal-Tools\\ (Single-Tool)\end{tabular} & \begin{tabular}[c]{@{}c@{}}Nexus\\ Raven\end{tabular} & \begin{tabular}[c]{@{}c@{}}Func\\ Name\end{tabular} & \begin{tabular}[c]{@{}c@{}}Func.+\\ Args\end{tabular} \\
\midrule
Qwen2-7B-Instruct & 81.55 $|$ 60.62 & 95.65 $|$ 49.50 & 71.59 $|$ 48.11 & 93.88 $|$ 77.51 & 87.05 $|$ 63.47 & 85.94 & 59.84 \\
\rowcolor[HTML]{EFEFEF} Hammer-7B & 93.48 $|$ 85.79 & 82.91 $|$ 66.40 & 82.31 $|$ 59.86 & 97.44 $|$ 91.66 & 92.46 $|$ 77.35 & 89.72 & 76.21 \\
Qwen1.5-4B-Chat & 55.33 $|$ 59.78 & 46.74 $|$ 38.48 & 35.41 $|$ 16.98 & 48.44 $|$ 62.32 & 29.03 $|$ 33.70 & 42.99 & 42.25 \\
\rowcolor[HTML]{EFEFEF} Hammer-4B & 91.65 $|$ 81.46 & 77.59 $|$ 61.01 & 85.09 $|$ 56.96 & 96.42 $|$ 92.45 & 81.73 $|$ 64.89 & 86.50 & 71.35 \\
Qwen2-1.5B-instruct & 74.63 $|$ 63.55 & 57.69 $|$ 33.62 & 65.76          $|$ 45.25 & 82.08 $|$ 75.49 & 70.62 $|$ 45.46 & 70.16 & 52.67 \\
\rowcolor[HTML]{EFEFEF} Hammer-1.5B & 82.13 $|$ 72.30 & 79.82 $|$ 59.71 & 80.90 $|$ 53.48 & 95.59 $|$ 88.65 & 79.87 $|$ 56.88 & 83.66 & 66.20 \\
\midrule
xLAM-7B-fc (FC) & 90.05 $|$ 80.69 & 72.49 $|$ 64.24 & 67.26 $|$ 58.96 & 78.97 $|$ 76.87 & 54.09 $|$ 57.50 & 72.57 & 67.65 \\
Deepseek-Coder-7B-Instruct & 51.42 $|$ 56.70 & 35.51 $|$ 39.64 & 11.58 $|$ 20.08 & 50.00 $|$ 65.47 & 26.89 $|$ 46.47 & 35.08 & 45.67 \\
\rowcolor[HTML]{EFEFEF} Deepseek-Coder-7B-Hammer & 83.47 $|$ 75.18 & 69.17 $|$ 60.04 & 83.77 $|$ 62.95 & 96.95 $|$ 93.20 & 93.75 $|$ 83.35 & 85.42 & 74.94 \\
xLAM-1.3B-fc (FC) & 94.86 $|$ 83.70 & 91.80 $|$ 64.32 & 64.86 $|$ 50.58 & 90.74 $|$ 80.43 & 64.43 $|$ 54.80 & 81.34 & 66.77 \\
Deepseek-Coder-1.3B-Instruct & 35.23 $|$ 38.42 & 20.41 $|$ 24.75 & 10.68 $|$ 06.06 & 16.46 $|$ 21.52 & 4.34 $|$ 8.15 & 17.42 & 19.78 \\
\rowcolor[HTML]{EFEFEF} Deepseek-Coder-1.3B-Hammer & 85.51 $|$ 77.22 & 77.68 $|$ 65.97 & 82.26 $|$ 57.68 & 95.74 $|$ 88.98 & 81.37 $|$ 64.68 & 84.51 & 70.91 \\
\bottomrule
\end{tabular}
\end{table}

\subsection{Ablation on Different Masking Ratio} 

To further investigate the impact of various function masking ratios on model performance, we designed an ablation study focused on the masking ratio. We systematically applied different masking ratios while fine-tuning the Qwen2-1.5B model on the Seal-Tools training dataset for one epoch. Subsequently, we evaluated the performance of the models trained with different masking ratios on the test sets of both Seal-Tools and API-Bank. This allowed us to observe and analyze the performance across both same-task and cross-task scenarios. 

\begin{figure*}[ht]
    \centering
    \begin{subfigure}{0.48\linewidth}
        \centering
        \includegraphics[width=2.9in]{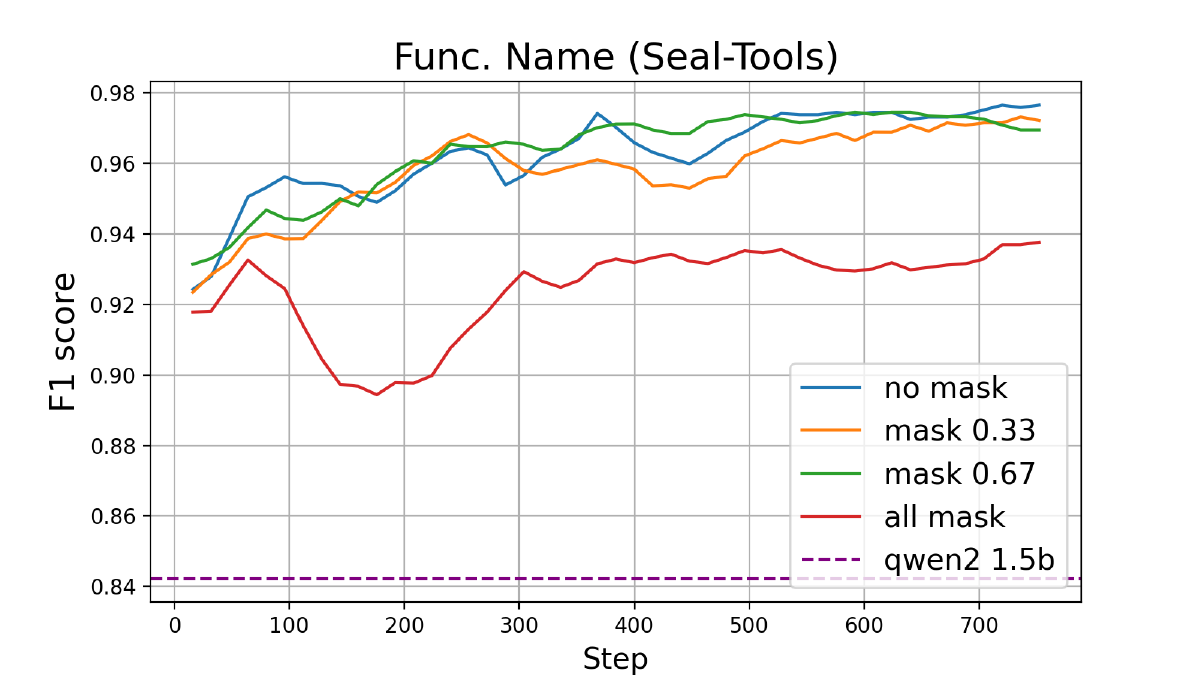}
    \end{subfigure}
    \begin{subfigure}{0.48\linewidth}
        \centering
        \includegraphics[width=2.9in]{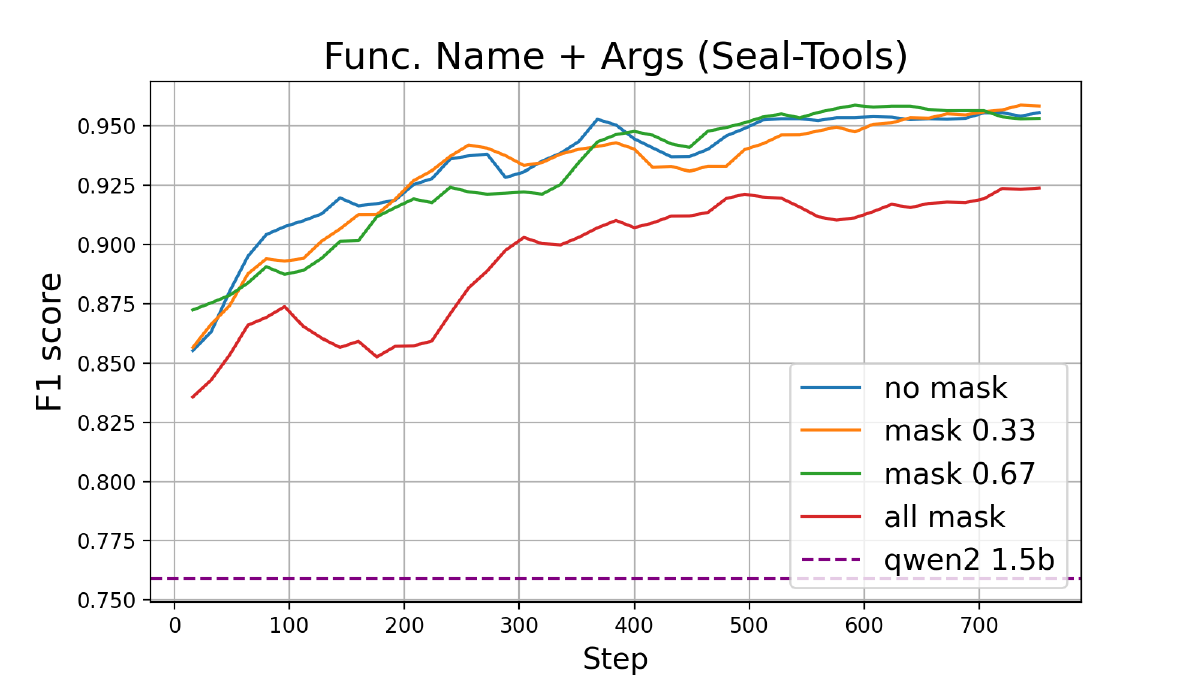}
    \end{subfigure}
    \begin{subfigure}{0.48\linewidth}
        \centering
        \includegraphics[width=2.9in]{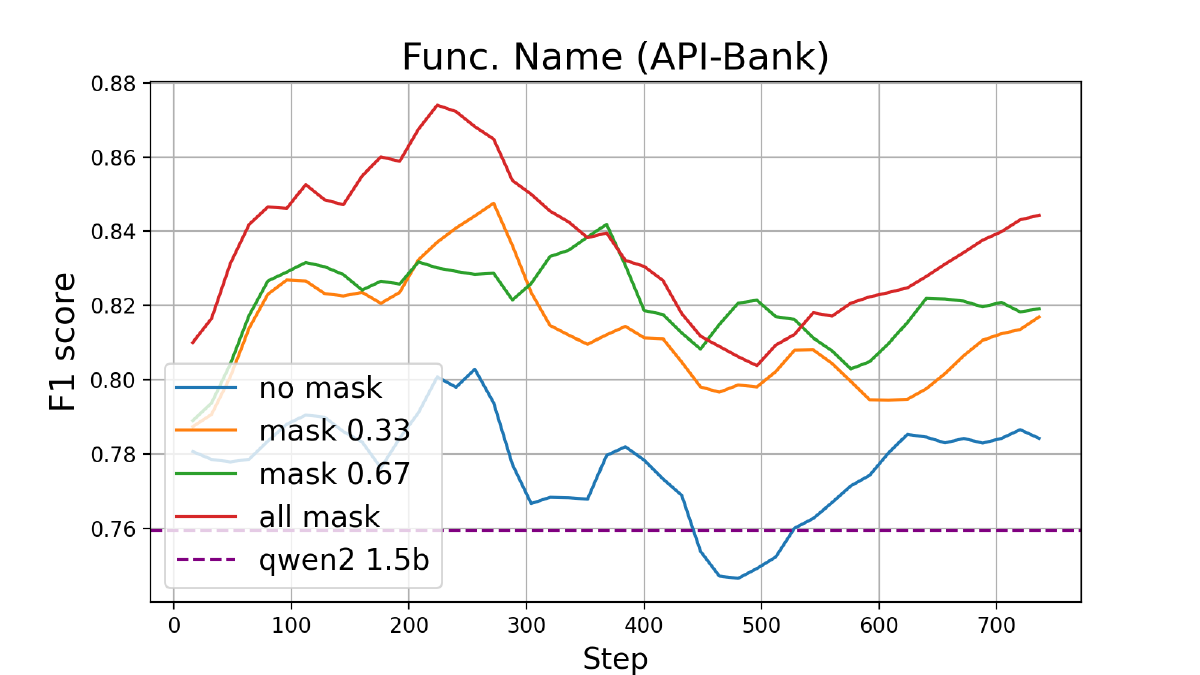}
    \end{subfigure}
    \begin{subfigure}{0.48\linewidth}
        \centering
        \includegraphics[width=2.9in]{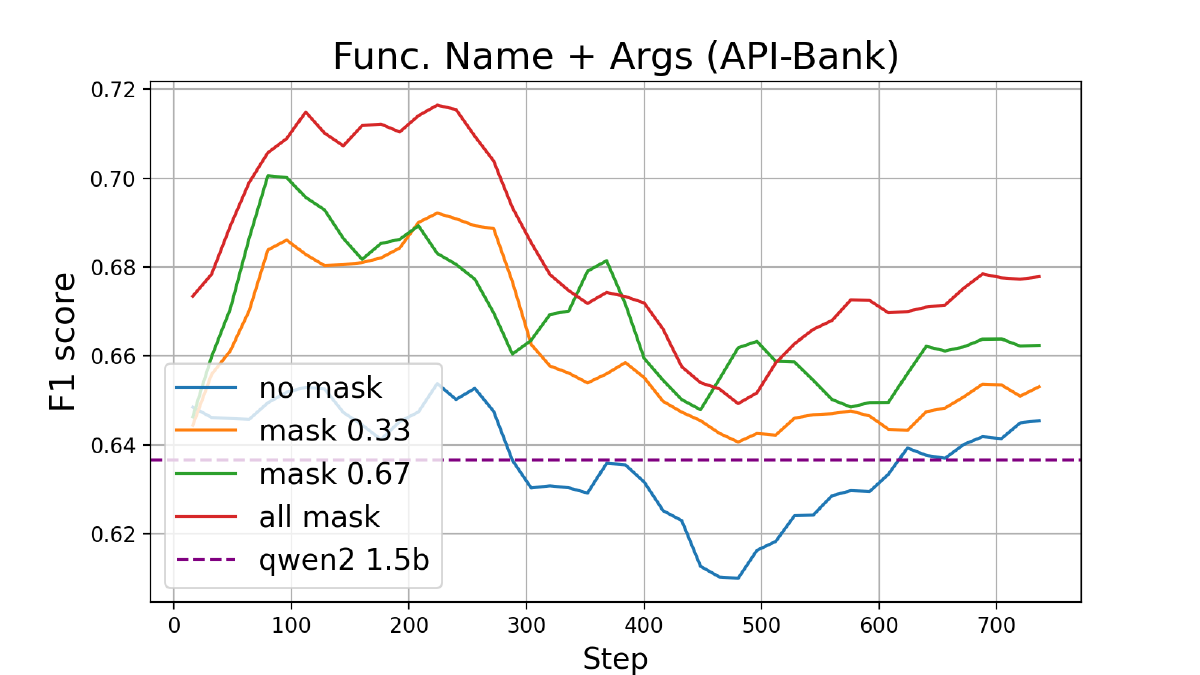}
    \end{subfigure}
    \caption{\normalsize An ablation to evaluate the impact of different masking ratios. For instance, ``mask 0.33'' denotes that 33\% of the instances in the training batch are masked, while others remain unaltered.} 
    \label{fig:abl-masking-ratios}
\end{figure*}

Based on the results presented in \autoref{fig:abl-masking-ratios}, we observe that an excessively large mask ratio can impede the model's learning speed within the same task scenario, i.e. the test on Seal-Tools. Conversely, the testing results on API-Bank indicate that a larger mask ratio facilitates better generalization of the model across different scenarios. This observation aligns with our previous insights, suggesting that, in the absence of masking, the model may overfit to the training data during fine-tuning, negatively impacting its performance in novel task environments. By enforcing a focus on more flexible description content, function masking can mitigate this overfitting to some extent, thereby enhancing cross-scenario generalization performance.

\subsection{Ablation on Different Proportions of Irrelevance-Augmented Data} 
\label{sec:ablation-irrelevance}

In Section 4.1, we discussed the trade-off between the model's performance in irrelevance detection and function-calling tasks, which motivated the design of the irrelevance-augmented dataset. To further explore the relationship between these two aspects, we conducted an ablation study on the data ratio of irrelevance-augmented data compared to the original xlam-function-calling data during training. In this ablation experiment, we sampled a total of 10,000 instances from both datasets at varying data proportions to fine-tune the Qwen2-1.5B-Instruct model and then exam on the BFCL testset, observing the changes in the model's irrelevance detection and function-calling capabilities across different ratios. 

\begin{figure*}[ht!]
    \centering
    \begin{subfigure}{0.32\linewidth}
        \centering
        \includegraphics[width=1.9in]{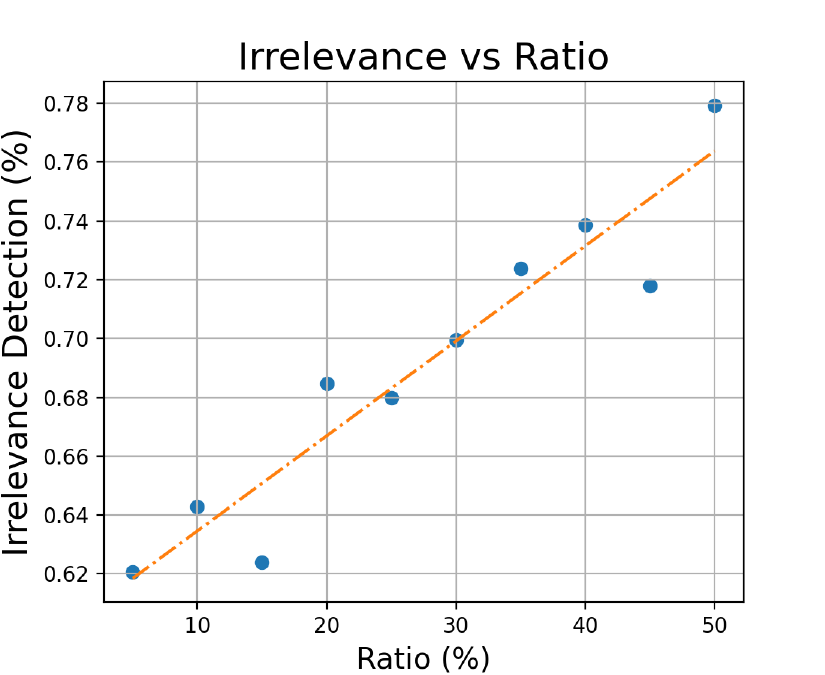}
    \end{subfigure}
    \begin{subfigure}{0.32\linewidth}
        \centering
        \includegraphics[width=1.9in]{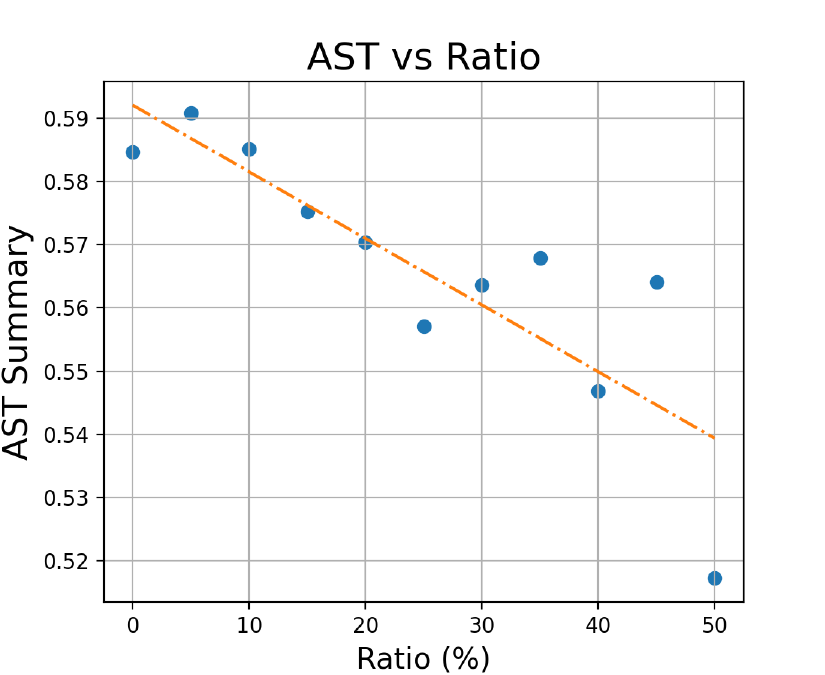}
    \end{subfigure}
    \begin{subfigure}{0.32\linewidth}
        \centering
        \includegraphics[width=1.9in]{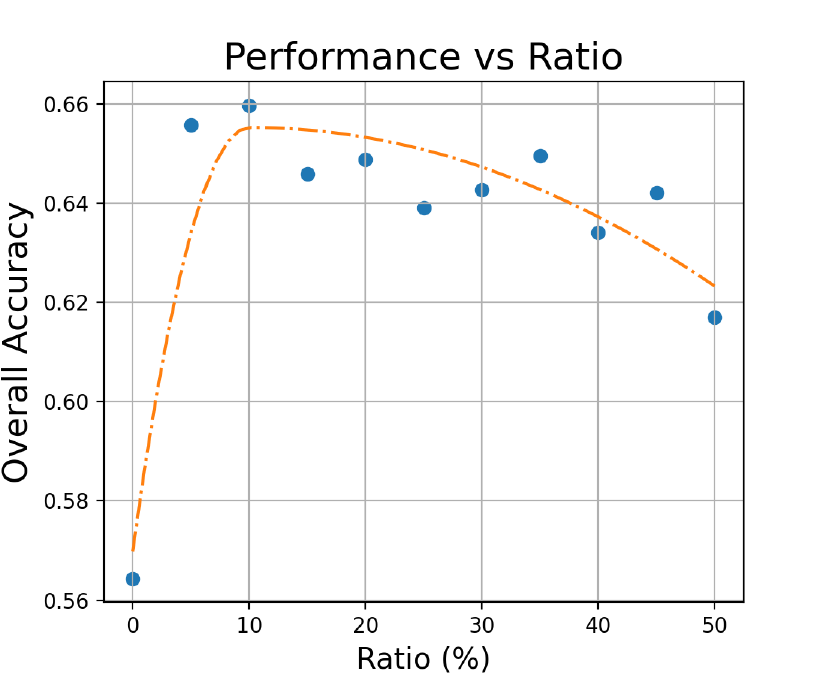}
    \end{subfigure}
    \caption{\normalsize Ablation on different proportions of irrelevance-augmented data applied, e.g. ratio=30\% means 30 percent of the training data is sampled from the irrelevance-augmented dataset with other 70 percent sampled from xlam-function-calling dataset.} 
    \label{fig:abl-irrelevance-ratios}
\end{figure*}

As illustrated in the first two panels of \autoref{fig:abl-irrelevance-ratios}, the variation in the proportion of irrelevance-augmented data reveals an inverse relationship between the model's performance in irrelevance detection and its function-calling capabilities. This finding underscores the importance of balancing the trade-off between these two aspects. Furthermore, the final panel of \autoref{fig:abl-irrelevance-ratios} indicates that, within our experimental settings, the Hammer model achieves optimal overall performance when the proportion of irrelevance-augmented data is approximately 10\%. This insight guides us in establishing the target size for the irrelevance-augmented dataset, i.e., 7.5k. It is essential to note that this proportion may require adjustment depending on the underlying model and training dataset; thus, the ratios presented herein are intended as a reference only.

\section{Conclusion}

In conclusion, our exploration of function-calling models reveals significant challenges related to performance inconsistency across different benchmarks, primarily driven by misleading from specific naming conventions. By introducing the Hammer family of models, we provide a robust solution that enhances generalization capabilities through a carefully constructed augmented dataset and innovative function masking techniques. The superior performance of Hammer on a variety of benchmarks demonstrates its potential for practical application in real-world scenarios. 

\newpage

\bibliographystyle{unsrtnat}
\bibliography{main}

\newpage

\appendix
\section{Extra Experimental Results}
\label{apx:extra-exp}

\begin{table}[h]
\caption{Full evaluation of the BFCL leaderboard \citep{berkeley-function-calling-leaderboard}. (as of date 09/20/2024.)}
\label{tab:full-bfcl}
\resizebox{\linewidth}{!}{%
\begin{tabular}{ccccccccccccc}
\toprule
                       &                                                &                               & \multicolumn{4}{c}{AST Category}                                                           & \multicolumn{4}{c}{Exec Category}                                                          &                               &                             \\ \cline{4-11}
\multirow{-2}{*}{Rank} & \multirow{-2}{*}{Model}                        & \multirow{-2}{*}{Overall Acc} & simple & Multiple & Parallel & \begin{tabular}[c]{@{}c@{}}Parallel\\ Multiple\end{tabular} & Simple & Multiple & Parallel & \begin{tabular}[c]{@{}c@{}}Parallel\\ Multiple\end{tabular} & \multirow{-2}{*}{Irrelevance} & \multirow{-2}{*}{Relevance} \\ \midrule
1                      & GPT-4-0125-Preview (Prompt)                    & 85.79                         & 78.82  & 88.44    & 91.00    & 83.75                                                       & 99.00  & 96.00    & 82.00    & 80.00                                                       & 61.35                         & 97.56                       \\
2                      & GPT-4-1106-Preview (Prompt)                    & 85.00                         & 78.75  & 89.12    & 94.12    & 83.25                                                       & 99.00  & 96.00    & 82.00    & 72.50                                                       & 64.98                         & 90.24                       \\
3                      & GPT-4-0613 (Prompt)                            & 84.74                         & 78.76  & 85.46    & 91.75    & 82.67                                                       & 98.29  & 96.00    & 86.00    & 70.00                                                       & 75.57                         & 82.93                       \\
\rowcolor[HTML]{EFEFEF} 
                       & Hammer-7b (FC)                                 & 83.92                         & 69.31  & 82.52    & 78.88    & 84.08                                                       & 91.86  & 94.00    & 88.00    & 85.00                                                       & 72.87                         & 92.68                       \\
4                      & GPT-4-turbo-2024-04-09 (Prompt)                & 83.89                         & 80.47  & 88.81    & 88.12    & 84.25                                                       & 99.00  & 96.00    & 80.00    & 77.50                                                       & 61.82                         & 82.93                       \\
5                      & GPT-4o-mini-2024-07-18 (Prompt)                & 83.35                         & 75.88  & 81.64    & 85.12    & 79.42                                                       & 98.29  & 94.00    & 82.00    & 77.50                                                       & 79.20                         & 80.49                       \\
6                      & GPT-4o-2024-05-13 (Prompt)                     & 83.13                         & 76.18  & 86.01    & 92.12    & 81.00                                                       & 98.00  & 94.00    & 76.00    & 72.50                                                       & 77.44                         & 78.05                       \\
7                      & Functionary-Medium-v3.1 (FC)                   & 82.55                         & 74.34  & 87.59    & 81.62    & 80.67                                                       & 98.29  & 94.00    & 90.00    & 75.00                                                       & 73.23                         & 70.73                       \\
8                      & GPT-4-1106-Preview (FC)                        & 81.78                         & 69.32  & 84.19    & 86.38    & 71.92                                                       & 95.43  & 94.00    & 86.00    & 75.00                                                       & 72.70                         & 82.93                       \\
9                      & Meta-Llama-3-70B-Instruct (Prompt)             & 81.59                         & 72.87  & 85.91    & 84.00    & 77.83                                                       & 94.14  & 94.00    & 84.00    & 80.00                                                       & 50.47                         & 92.68                       \\
10                     & Claude-3-Opus-20240229 (Prompt)                & 80.88                         & 76.65  & 87.47    & 78.38    & 75.17                                                       & 98.57  & 94.00    & 82.00    & 75.00                                                       & 56.15                         & 85.37                       \\
11                     & GPT-4-0125-Preview (FC)                        & 80.87                         & 68.76  & 84.95    & 80.38    & 74.00                                                       & 84.21  & 94.00    & 88.00    & 75.00                                                       & 74.03                         & 85.37                       \\
12                     & Nemotron-4-340b-instruct (Prompt)              & 80.23                         & 68.51  & 80.38    & 78.62    & 79.17                                                       & 86.00  & 90.00    & 80.00    & 77.50                                                       & 84.10                         & 78.05                       \\
13                     & Functionary-Small-v3.1 (FC)                    & 80.21                         & 72.70  & 83.31    & 85.62    & 72.92                                                       & 87.79  & 90.00    & 86.00    & 70.00                                                       & 68.36                         & 85.37                       \\
14                     & mistral-large-2407 (FC Any)                    & 79.66                         & 81.01  & 87.42    & 90.50    & 83.50                                                       & 98.29  & 92.00    & 86.00    & 77.50                                                       & 0.34                          & 100.00                      \\
15                     & GPT-4o-2024-05-13 (FC)                         & 79.55                         & 70.40  & 82.33    & 89.00    & 76.08                                                       & 88.93  & 84.00    & 88.00    & 72.50                                                       & 73.50                         & 70.73                       \\
16                     & xLAM-7b-fc-r (FC)                              & 79.41                         & 70.28  & 78.18    & 74.12    & 68.50                                                       & 94.21  & 88.00    & 88.00    & 72.50                                                       & 79.76                         & 80.49                       \\
17                     & GPT-4o-mini-2024-07-18 (FC)                    & 79.25                         & 67.83  & 80.16    & 85.38    & 77.17                                                       & 83.21  & 92.00    & 82.00    & 70.00                                                       & 71.83                         & 82.93                       \\
18                     & Open-Mixtral-8x22b (Prompt)                    & 79.14                         & 73.47  & 76.14    & 79.12    & 73.67                                                       & 91.86  & 96.00    & 84.00    & 75.00                                                       & 71.42                         & 70.73                       \\
19                     & Gorilla-OpenFunctions-v2 (FC)                  & 79.10                         & 70.81  & 79.47    & 75.75    & 66.67                                                       & 95.86  & 96.00    & 78.00    & 70.00                                                       & 73.13                         & 85.37                       \\
20                     & GPT-4-turbo-2024-04-09 (FC)                    & 79.09                         & 64.21  & 82.72    & 82.50    & 75.75                                                       & 88.71  & 88.00    & 86.00    & 72.50                                                       & 79.79                         & 70.73                       \\
21                     & Functionary-Small-v3.2 (FC)                    & 78.96                         & 69.50  & 81.50    & 80.12    & 73.50                                                       & 90.64  & 88.00    & 86.00    & 67.50                                                       & 72.32                         & 80.49                       \\
22                     & GPT-4o-2024-08-06 (FC)                         & 78.87                         & 70.71  & 80.97    & 83.25    & 75.58                                                       & 85.36  & 90.00    & 84.00    & 72.50                                                       & 82.91                         & 63.41                       \\
23                     & mistral-large-2407 (FC Auto)                   & 78.78                         & 68.28  & 86.44    & 90.25    & 83.50                                                       & 76.86  & 92.00    & 86.00    & 77.50                                                       & 48.93                         & 78.05                       \\
24                     & Claude-3-Sonnet-20240229 (Prompt)              & 77.92                         & 71.80  & 85.26    & 82.75    & 73.92                                                       & 96.14  & 90.00    & 84.00    & 77.50                                                       & 30.01                         & 87.80                       \\
25                     & FireFunction-v2 (FC)                           & 77.45                         & 74.11  & 81.49    & 73.62    & 67.58                                                       & 94.43  & 88.00    & 82.00    & 72.50                                                       & 52.94                         & 87.80                       \\
26                     & Granite-20b-FunctionCalling (FC)               & 76.63                         & 65.27  & 73.05    & 60.75    & 67.83                                                       & 85.36  & 90.00    & 84.00    & 72.50                                                       & 72.43                         & 95.12                       \\
27                     & Open-Mistral-Nemo-2407 (Prompt)                & 76.31                         & 72.89  & 81.37    & 81.50    & 73.75                                                       & 92.50  & 94.00    & 86.00    & 80.00                                                       & 13.25                         & 87.80                       \\
28                     & Claude-3.5-Sonnet-20240620 (Prompt)            & 76.29                         & 76.98  & 80.27    & 72.62    & 65.33                                                       & 98.50  & 92.00    & 70.00    & 72.50                                                       & 83.46                         & 51.22                       \\
\rowcolor[HTML]{EFEFEF} 
                       & Hammer-4b (FC)                                 & 76.05                         & 62.58  & 77.72    & 69.12    & 68.92                                                       & 67.79  & 92.00    & 86.00    & 77.50                                                       & 68.66                         & 90.24                       \\
29                     & GPT-3.5-Turbo-0125 (FC)                        & 75.41                         & 69.79  & 83.58    & 71.88    & 68.83                                                       & 95.14  & 88.00    & 86.00    & 57.50                                                       & 35.83                         & 97.56                       \\
30                     & Open-Mistral-Nemo-2407 (FC Auto)               & 74.97                         & 64.57  & 79.99    & 80.25    & 74.00                                                       & 91.36  & 86.00    & 86.00    & 62.50                                                       & 59.14                         & 65.85                       \\
31                     & xLAM-1b-fc-r (FC)                              & 74.90                         & 64.49  & 73.06    & 64.00    & 67.92                                                       & 79.21  & 88.00    & 86.00    & 70.00                                                       & 61.21                         & 95.12                       \\
32                     & Hermes-2-Pro-Llama-3-70B (FC)                  & 74.78                         & 66.29  & 73.49    & 70.25    & 78.33                                                       & 80.64  & 88.00    & 84.00    & 72.50                                                       & 53.80                         & 80.49                       \\
33                     & Gemini-1.5-Pro-Preview-0514 (FC)               & 74.75                         & 56.15  & 78.89    & 82.38    & 65.50                                                       & 75.71  & 88.00    & 84.00    & 75.00                                                       & 83.31                         & 58.54                       \\
34                     & Claude-2.1 (Prompt)                            & 74.57                         & 68.21  & 78.08    & 74.12    & 66.17                                                       & 94.64  & 88.00    & 64.00    & 62.50                                                       & 74.36                         & 75.61                       \\
35                     & Gemini-1.5-Pro-Preview-0409 (FC)               & 74.56                         & 55.08  & 79.43    & 83.12    & 64.75                                                       & 76.00  & 88.00    & 80.00    & 72.50                                                       & 83.27                         & 63.41                       \\
36                     & GPT-4o-2024-08-06 (Prompt)                     & 74.12                         & 65.76  & 76.86    & 72.12    & 71.67                                                       & 70.57  & 88.00    & 78.00    & 75.00                                                       & 89.56                         & 53.66                       \\
37                     & Command-R-Plus (Prompt) (Original)             & 74.11                         & 68.14  & 78.13    & 77.50    & 62.17                                                       & 91.29  & 86.00    & 78.00    & 55.00                                                       & 69.31                         & 75.61                       \\
38                     & Open-Mistral-Nemo-2407 (FC Any)                & 73.12                         & 67.98  & 82.46    & 77.38    & 76.08                                                       & 92.07  & 86.00    & 86.00    & 62.50                                                       & 0.72                          & 100.00                      \\
\rowcolor[HTML]{EFEFEF} 
                       & Hammer-1.5b (FC)                               & 73.04                         & 62.34  & 72.84    & 58.75    & 68.17                                                       & 49.93  & 92.00    & 84.00    & 77.50                                                       & 72.18                         & 92.68                       \\
39                     & Mistral-Medium-2312 (Prompt)                   & 72.19                         & 63.77  & 80.22    & 69.12    & 59.25                                                       & 93.43  & 88.00    & 70.00    & 57.50                                                       & 84.54                         & 56.10                       \\
40                     & Command-R-Plus (FC) (Original)                 & 72.04                         & 64.25  & 72.45    & 66.25    & 62.33                                                       & 89.14  & 86.00    & 82.00    & 52.50                                                       & 52.75                         & 92.68                       \\
41                     & Gemini-1.5-Flash-Preview-0514 (FC)             & 70.75                         & 65.80  & 83.26    & 63.87    & 63.50                                                       & 57.93  & 86.00    & 74.00    & 75.00                                                       & 74.69                         & 63.41                       \\
42                     & DBRX-Instruct (Prompt)                         & 69.55                         & 69.97  & 80.35    & 66.88    & 51.50                                                       & 90.50  & 86.00    & 60.00    & 62.50                                                       & 44.86                         & 82.93                       \\
43                     & Claude-3.5-Sonnet-20240620 (FC)                & 68.88                         & 73.95  & 82.09    & 65.38    & 62.75                                                       & 95.36  & 86.00    & 44.00    & 40.00                                                       & 75.91                         & 63.41                       \\
44                     & GPT-3.5-Turbo-0125 (Prompting)                 & 66.19                         & 59.01  & 67.74    & 65.25    & 48.58                                                       & 44.50  & 86.00    & 78.00    & 55.00                                                       & 69.97                         & 87.80                       \\
45                     & Hermes-2-Pro-Llama-3-8B (FC)                   & 66.18                         & 62.32  & 74.96    & 61.62    & 57.83                                                       & 68.71  & 90.00    & 80.00    & 57.50                                                       & 55.16                         & 53.66                       \\
46                     & Hermes-2-Pro-Mistral-7B (FC)                   & 65.44                         & 60.98  & 71.49    & 60.38    & 50.42                                                       & 60.50  & 90.00    & 84.00    & 62.50                                                       & 38.55                         & 75.61                       \\
47                     & Hermes-2-Theta-Llama-3-8B (FC)                 & 64.83                         & 58.53  & 67.82    & 59.62    & 58.33                                                       & 69.14  & 88.00    & 78.00    & 55.00                                                       & 62.66                         & 51.22                       \\
48                     & Meta-Llama-3-8B-Instruct (Prompt)              & 62.70                         & 58.53  & 70.26    & 53.50    & 53.25                                                       & 84.50  & 88.00    & 68.00    & 50.00                                                       & 22.88                         & 78.05                       \\
49                     & Claude-3-Opus-20240229 (FC tools-2024-04-04)   & 61.89                         & 69.41  & 79.95    & 39.38    & 27.92                                                       & 84.64  & 86.00    & 52.00    & 30.00                                                       & 76.40                         & 73.17                       \\
50                     & Open-Mixtral-8x7b (Prompt)                     & 60.82                         & 61.49  & 70.70    & 47.12    & 36.83                                                       & 71.86  & 74.00    & 56.00    & 52.50                                                       & 71.84                         & 65.85                       \\
51                     & Claude-3-Haiku-20240307 (Prompt)               & 60.34                         & 74.64  & 84.49    & 51.88    & 45.17                                                       & 89.43  & 94.00    & 32.00    & 27.50                                                       & 18.90                         & 85.37                       \\
52                     & Open-Mixtral-8x22b (FC Any)                    & 58.89                         & 73.23  & 85.42    & 10.75    & 63.08                                                       & 92.57  & 92.00    & 24.00    & 47.50                                                       & 0.34                          & 100.00                      \\
53                     & Open-Mixtral-8x22b (FC Auto)                   & 58.37                         & 59.75  & 82.75    & 10.50    & 62.33                                                       & 77.79  & 92.00    & 24.00    & 45.00                                                       & 44.20                         & 85.37                       \\
54                     & Gemini-1.0-Pro-001 (FC)                        & 57.81                         & 64.90  & 79.40    & 38.12    & 22.25                                                       & 86.14  & 84.00    & 58.00    & 5.00                                                        & 67.13                         & 73.17                       \\
55                     & Mistral-small-2402 (FC Auto)                   & 55.36                         & 51.90  & 82.00    & 15.62    & 34.33                                                       & 87.57  & 90.00    & 14.00    & 20.00                                                       & 77.67                         & 80.49                       \\
56                     & Mistral-small-2402 (FC Any)                    & 52.45                         & 65.89  & 84.78    & 15.88    & 36.42                                                       & 94.71  & 90.00    & 14.00    & 22.50                                                       & 0.34                          & 100.00                      \\
57                     & FireFunction-v1 (FC)                           & 48.11                         & 72.25  & 80.37    & 0.00     & 0.00                                                        & 84.79  & 80.00    & 0.00     & 0.00                                                        & 68.55                         & 95.12                       \\
58                     & Claude-3-Sonnet-20240229 (FC tools-2024-04-04) & 47.97                         & 63.79  & 78.37    & 8.25     & 3.33                                                        & 78.50  & 90.00    & 0.00     & 0.00                                                        & 59.89                         & 97.56                       \\
59                     & Claude-instant-1.2 (Prompt)                    & 47.95                         & 54.50  & 55.81    & 37.75    & 32.42                                                       & 57.50  & 72.00    & 38.00    & 15.00                                                       & 70.21                         & 46.34                       \\
60                     & Claude-3-Haiku-20240307 (FC tools-2024-04-04)  & 47.03                         & 72.74  & 78.95    & 1.00     & 2.33                                                        & 90.64  & 92.00    & 6.00     & 0.00                                                        & 29.08                         & 97.56                       \\
61                     & GPT-4-0613 (FC)                                & 45.61                         & 56.33  & 86.36    & 0.00     & 0.00                                                        & 69.21  & 90.00    & 0.00     & 0.00                                                        & 80.99                         & 73.17                       \\
62                     & Snowflake/snowflake-arctic-instruct (Prompt)   & 42.46                         & 34.97  & 31.79    & 42.00    & 38.33                                                       & 33.29  & 28.00    & 60.00    & 40.00                                                       & 65.01                         & 51.22                       \\
63                     & mistral-large-2407 (Prompt)                    & 27.87                         & 18.08  & 43.11    & 33.38    & 23.17                                                       & 8.71   & 30.00    & 18.00    & 5.00                                                        & 40.70                         & 58.54                       \\
64                     & Mistral-Small-2402 (Prompt)                    & 24.44                         & 7.83   & 38.97    & 17.25    & 8.92                                                        & 8.14   & 12.00    & 12.00    & 0.00                                                        & 83.22                         & 56.10                       \\
65                     & Mistral-tiny-2312 (Prompt)                     & 21.17                         & 21.11  & 25.92    & 9.75     & 3.50                                                        & 19.64  & 8.00     & 12.00    & 0.00                                                        & 92.23                         & 19.51                       \\
66                     & Deepseek-v1.5 (Prompt)                         & 11.18                         & 4.07   & 0.00     & 1.00     & 2.83                                                        & 0.00   & 0.00     & 4.00     & 0.00                                                        & 99.89                         & 0.00                        \\
67                     & Gemma-7b-it (Prompt)                           & 10.30                         & 2.40   & 0.99     & 0.50     & 0.50                                                        & 1.71   & 0.00     & 0.00     & 0.00                                                        & 96.95                         & 0.00                        \\
68                     & Hermes-2-Theta-Llama-3-70B (FC)                & 10.00                         & 0.00   & 0.00     & 0.00     & 0.00                                                        & 0.00   & 0.00     & 0.00     & 0.00                                                        & 100.00                        & 0.00                        \\ \bottomrule
\end{tabular}
}
\end{table}

\begin{table}[ht]
\caption{AST Evaluation for Hammers and different base models on BFCL.}
\tiny
\begin{tabular}{lcccccccc}
\toprule
Overall Acc & Model & AST Summary & Simple & Multiple & Parallel & Parallel Multiple & Irrelevance & Relevance \\
\midrule
72.79 & Qwen2-7B-Instruct & 69.47 & 68.75 & 81.88        & 60.75 & 66.50 & 61.31 & 97.56 \\
\rowcolor[HTML]{EFEFEF} 
80.06 & Hammer-7B & 78.70 & 69.31 & 82.52 & 78.88 & 84.08 & 72.87 & 92.68 \\
32.92 & Qwen1.5-4B-Chat & 25.43 & 24.60 & 32.99 & 22.12 & 22.00 & 66.56 & 29.27 \\
\rowcolor[HTML]{EFEFEF} 72.87 & Hammer-4B & 69.59 & 62.58 & 77.72 & 69.12 & 68.92 & 68.66 & 90.24 \\
46.90 & Qwen2-1.5B-Instruct & 41.44 & 50.77 & 61.80 & 19.38 & 33.83 & 22.91 & 92.68 \\
\rowcolor[HTML]{EFEFEF} 71.16 & Hammer-1.5B & 65.52 & 62.34 & 72.84 & 58.75 & 68.17 & 72.18 & 92.68 \\
\midrule
75.22 & xLAM-7B-fc (FC) & 72.77 & 70.28 & 78.18 & 74.12 & 68.50 & 79.76 & 80.49 \\
17.65 & Deepseek-Coder-7B-Instruct & 1.60 & 3.53 & 0.05 & 0.25 & 2.58 & 99.51 & 0.00 \\
\rowcolor[HTML]{EFEFEF} 79.09 & Deepseek-Coder-7B-Instruct-Hammer & 76.84 & 71.03 & 84.51 & 77.00 & 74.83 & 67.14 & 100.00 \\
70.96 & xLAM-1.3B-fc (FC) & 67.37 & 64.49 & 73.06 & 64.00 & 67.92 & 61.21 & 95.12 \\
16.81 & Deepseek-Coder-1.3B-Instruct & 0.21 & 0.83 & 0.00 & 0.00 & 0.00 & 100.00 & 0.00 \\
\rowcolor[HTML]{EFEFEF} 69.71 & Deepseek-Coder-1.3B-Instruct-Hammer & 67.52 & 65.47 & 74.71 & 60.88 & 69.00 & 57.93 & 90.24 \\
\bottomrule
\end{tabular}
\end{table}

\section{Different Types of Function-calling Tasks}
\label{apx:func-type}

\textbf{Simple:} This query style includes straightforward scenarios where a single function call is made based on the user’s input with a single provided JSON format API description.

\textbf{Multiple:} In this style, user queries could be answered by one of several function calls. The challenge lies in selecting the most appropriate function from multiple provided APIs. It represents one of the most common real-world use cases.

\textbf{Parallel:} This query style requires executing multiple function calls simultaneously in response to a single user query, which may consist of one or more sentences but with only one API provided. 

\textbf{Parallel Multiple:} This query style combines the parallel and multiple categories, where multiple function and API documents are provided, and each function call might be invoked multiple times based on the query’s requirements.

\textbf{Irrelevance: } In this query style, no suitable function exists within the candidate options to fulfill users' intent, thus the model should have the ability to detect it and decline the task, rather than making incorrect attempts.

\begin{figure*}[h]
\centering
\centerline{\includegraphics[width=.99\columnwidth]{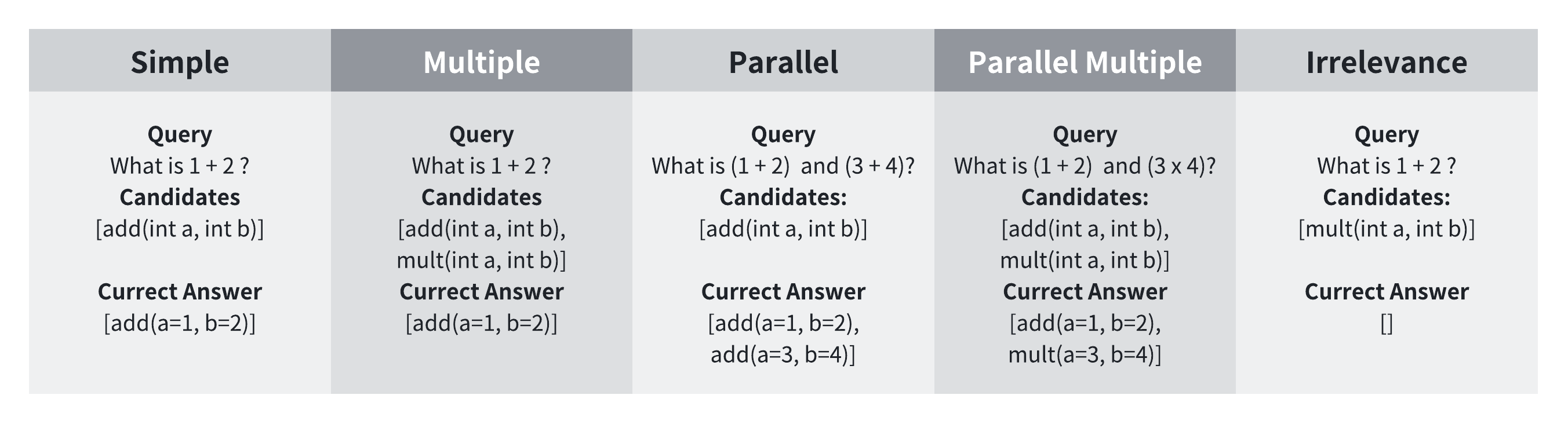}}
\caption{\normalsize Demonstration of different function-calling styles.}
\end{figure*}

\section{Example Input to Models with Function Masking}

The prompted inputs to models in our experiment are exampled as:

\begin{minted}{markdown}
[BEGIN OF TASK INSTRUCTION]
You are a tool calling assistant. In order to complete the user's 
request, you need to select one or more appropriate tools from the 
following tools and fill in the correct values for the tool parameters. 
Your specific tasks are:
1. Make one or more function/tool calls to meet the request based 
    on the question.
2. If none of the function can be used, point it out and refuse to 
    answer.
3. If the given question lacks the parameters required by  the function, 
    also point it out.
[END OF TASK INSTRUCTION]

[BEGIN OF AVAILABLE TOOLS]
[
    {
        "name": "LxOm64zLyg", 
        "description": "Gets hourly weather forecast information for 
            given geographical coordinates using the RapidAPI service.", 
        "parameters": {
            "TDpjPd": {
                "description": "The latitude of the geographical location.", 
                "type\": "int", 
                "default": 46.95828
            }, 
            "78th2U3lFj": {
                "description": "The longitude of the geographical location.", 
                "type": "int", 
                "default": 10.87152
            }
        }
    }, 
    {
        "name": "WoDdNSe7e7K5", 
        "description": "Fetches weather updates for a given city 
            using the RapidAPI Weather API.", 
        "parameters": {
            "LzZsvxUC": {
                "description": "The name of the city for which to 
                    retrieve weather information.", 
                "type": "str", 
                "default": "London"
            }
        }
    }, 
    {
        "name": "CBrCNmwOERb", 
        "description": "Fetches the hourly weather forecast for a 
            given location using the RapidAPI service.", 
        "parameters": {
            "TDEJ.ZwMt": {
                "description": "The name of the location for which 
                    to retrieve the hourly weather forecast.", 
                "type": "str", 
                "default": "Berlin"
            }
        }
    }, 
    {
        "name": "1YTQVXkwLY", 
        "description": "Returns an air quality forecast for a given 
            location.", 
        "parameters": {
            "2bkgDA": {
                "description": "The latitude of the location for 
                    which the air quality forecast is to be retrieved.", 
                "type": "int", 
                "default": "35.779"
            }, 
            "DQi.ReZ16": {
                "description": "The longitude of the location for 
                    which the air quality forecast is to be retrieved.", 
                "type": "int", 
                "default": "-78.638"
            }, 
            "hF.1": {
                "description": "The number of hours for which the 
                    forecast is to be retrieved (default is 72).", 
                "type": "int", 
                "default": "72"
            }
        }
    }
]

[END OF AVAILABLE TOOLS]

[BEGIN OF FORMAT INSTRUCTION]
The output MUST strictly adhere to the following JSON format, and 
NO other text MUST be included.\nThe example format is as follows. 
Please make sure the parameter type is correct. If no function 
call is needed, please directly output an empty list '[]'
```
[
    {
        "name": "func_name1", 
        "arguments": {
            "argument1": "value1", 
            "argument2": "value2"
        }
    },
... (more tool calls as required)
]
```
[END OF FORMAT INSTRUCTION]

[BEGIN OF QUERY]
What are the current weather conditions in Sydney?
[END OF QUERY]
\end{minted}

The corresponding output generated by models should be:

\begin{minted}{markdown}
```
[
    {
        "name": "WoDdNSe7e7K5", 
        "arguments": {
            "LzZsvxUC": "Sydney"
        }
    }
]
```
\end{minted}

% \newpage
% \bibliographystyle{unsrtnat}
% \bibliography{main}

% \newpage
% \input{appendix}

\end{document}